\pdfoutput=1

\documentclass[11pt]{article}

\usepackage{EMNLP2022}

\usepackage{times}
\usepackage{latexsym}

\usepackage[T1]{fontenc}

\usepackage[utf8]{inputenc}

\usepackage{microtype}

\usepackage{inconsolata}

\usepackage{url}
\usepackage{graphicx}
\usepackage{wrapfig}
\usepackage{amsmath}
\usepackage{amssymb}
\usepackage{color,xcolor,colortbl}
\usepackage{enumitem}
\usepackage{algorithm}
\usepackage{algorithmic}
\usepackage[font={small}]{caption}
\usepackage{bm,bbm}
\usepackage{booktabs}
\usepackage{mathtools}
\usepackage{array}
\usepackage{multirow}
\usepackage{soul}
\usepackage{subcaption}
\usepackage{pbox}
\usepackage{pifont}
\usepackage{arydshln}
\usepackage{comment}
\usepackage{array}

\newcommand\numberthis{\addtocounter{equation}{1}\tag{\theequation}}
\newcommand{\norm}[1]{\left\lVert#1\right\rVert}

\definecolor{darkblue}{rgb}{0.0, 0.0, 0.55}
\definecolor{midnightblue}{HTML}{191970}
\definecolor{darkgreen}{HTML}{006400}
\definecolor{red}{HTML}{ff0000}
\definecolor{gold}{HTML}{ffd700}
\definecolor{mediumvioletred}{HTML}{c71585}
\definecolor{aqua}{HTML}{00ffff}
\definecolor{fuchsia}{HTML}{ff00ff}
\definecolor{lightpink}{HTML}{ffb6c1}
\definecolor{dodgerblue}{HTML}{1e90ff}
\definecolor{deepskyblue}{HTML}{00bfff}
\definecolor{deeppink}{HTML}{ff1493}
\definecolor{orangered}{HTML}{ff4500}
\definecolor{mediumseagreen}{HTML}{3cb371}
\definecolor{saddlebrown}{HTML}{8b4513}

\definecolor{teal}{HTML}{469990}
\definecolor{green}{HTML}{3cb44b}
\definecolor{lime}{HTML}{bfef45}
\definecolor{mint}{HTML}{aaffc3}

\title{Toward Unifying Text Segmentation and Long Document Summarization}

\author{Sangwoo Cho,$^1$ Kaiqiang Song,$^1$ Xiaoyang Wang,$^1$ Fei Liu,$^2$ Dong Yu$^1$\\[0.6em]
$^1$Tencent AI Lab, Bellevue, WA\\
$^2$Department of Computer Science, Emory University, Atlanta, GA\\[0.6em]
\texttt{\{swcho, riversong, shawnxywang, dyu\}@global.tencent.com}\\ 
\texttt{fei.liu@emory.edu}
}

\begin{document}
\maketitle
\begin{abstract}

Text segmentation is important for signaling a document's structure.
Without segmenting a long document into topically coherent sections,
it is difficult for readers to comprehend the text, let alone find important information.
The problem is only exacerbated by a lack of segmentation in transcripts of audio/video recordings. 
In this paper, we explore the role that \emph{section segmentation} plays
in extractive summarization of written and spoken documents. 
Our approach learns robust sentence representations by performing summarization and segmentation simultaneously,
which is further enhanced by an optimization-based regularizer to promote selection of diverse summary sentences.
We conduct experiments on multiple datasets ranging from scientific articles to spoken transcripts to evaluate the model's performance.
Our findings suggest that the model can not only achieve state-of-the-art performance on publicly available benchmarks,
but demonstrate better cross-genre transferability when equipped with text segmentation.
We perform a series of analyses to quantify the impact of section segmentation on 
summarizing written and spoken documents of substantial length and complexity.

\end{abstract}

\section{Introduction}

One of the most effective ways to summarize a long document is to extract salient sentences~\cite{goldstein-etal-2000-multi}.
While abstractive strategies produce more condensed summaries,
they suffer from hallucinations and factual errors,
which pose a more difficult generation challenge~\cite{lebanoff-etal-2020-understanding,goyal-durrett-2021-annotating}.
In this study, we focus on extractive summarization of lengthy documents,
including both written documents and transcripts of spoken language.
Extractive summaries have the potential to be highlighted on their source materials to facilitate viewing,
e.g., Google's browser allows text extracts to be highlighted on the webpage via a shareable link~\cite{HighlightedLinks:2021}.

As a document grows in length, it becomes crucial to bring structure to it.
Examples include chapters, sections, paragraphs, headings and bulleted lists~\cite{power-etal-2003-document}.
All of them allow readers to find salient content buried within the document.
Particularly, having \emph{sections} is a differentiating factor between a long and a mid-length document.  
A long document such as a research article contains over 5,000 words~\cite{cohan-etal-2018-discourse}.
It is an order of magnitude longer than a mid-length document such as a news article~\cite{see-etal-2017-get}.
Writing a long document thus requires the author to meticulously organize the content into sections.
In this paper, we equip our summarizer with the ability to predict section boundaries 
and leverage this ability to improve long document summarization.

Importantly, sections are essential to both written and spoken documents.
A majority of summarization approaches concentrate on written documents, assuming the sections are given.
They exploit document structure by hierarchically building representations from words to sentences, 
then to larger sections and documents~\cite{xiao-carenini-2019-extractive,liu-lapata-2019-text,narayan-etal-2020-stepwise}.
It remains an open question as to whether spoken transcripts can be handled in a similar manner.
E.g., the transcript for a 1.5-hour video lecture contains >10,000 words.
There are no section boundaries.
Instead, the lecture content is loosely organized around a series of talking points.
Discourse cues, e.g., ``\emph{so next we need to...},'' have been shown to correlate with the underlying document structure~\cite{hearst-1997-text}.
We thus aim to leverage such cues to infer section boundaries, which help summarization of both spoken and written documents.

Our model learns robust sentence representations by performing the two tasks of extractive summarization and section segmentation simultaneously,
enhanced by an optimization-based framework to select important and diverse sentences.
It mimics what a human would do when identifying salient content from a lengthy document. 
Text segmentation was previously studied as a standalone problem~\cite{arnold-etal-2019-sector,xing-etal-2020-improving,lukasik-etal-2020-text}.
For example, \citet{koshorek-etal-2018-text} break Wikipedia articles into sections according to tables of contents. 
In this work, we enhance extractive summarization with a new addition of section segmentation. 
We train our model on written documents with known section boundaries, 
then adapt it to transcripts where such information is unavailable to exploit its transferability.
We observe that by predicting section boundaries, our model learns to not only encode salient content 
but also recognize document structure information. 

\begin{figure*}
\centering
\includegraphics[width=6in]{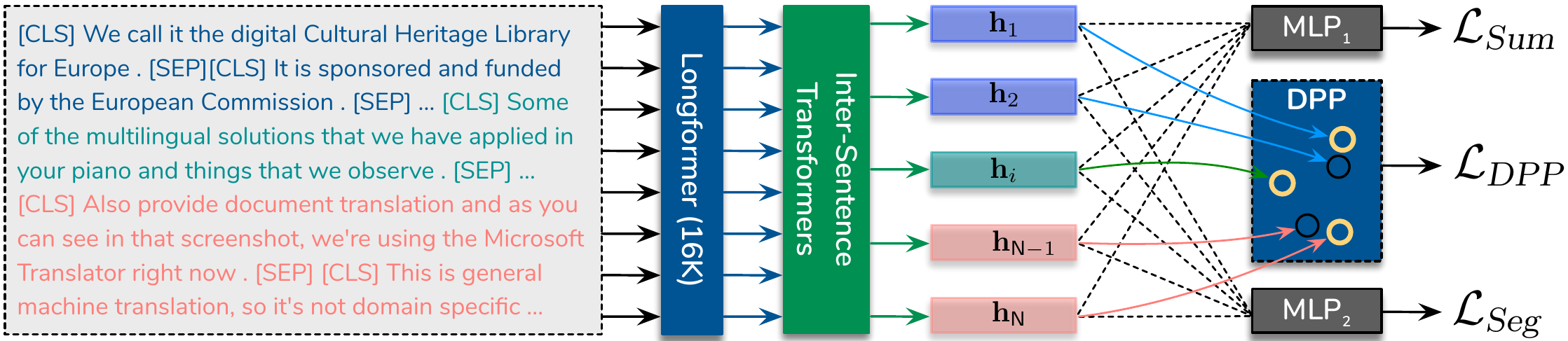}
\vspace{-0.05in}
\caption{An overview of our system named ``\textbf{\textsl{Lodoss}}.'' It builds effective sentence representations by combining two essential tasks of section segmentation and sentence extraction. We introduce a new regularizer $\mathcal{L}_{\scriptsize\mbox{DPP}}$ drawing on determinantal point processes to collectively measure the quality of a set of extracted sentences, ensuring they are informative and diverse. 
}
\label{fig:network}
\end{figure*}

Ensuring that a summary covers a broad range of important topics is pivotal.
A long document may discuss multiple topics. 
It is inadequate for a summary to have a narrow information focus and miss the important points of the document.
Crucially, we design a new regularizer drawing on learned sentence representations and determinantal point process \cite{Kulesza:2012,cho-etal-2019-improving}
to ensure a set of representative and diverse sentences
is selected for the summary.
We evaluate our proposed approach against strong summarization baselines 
and on multiple datasets ranging from scientific articles to lecture transcripts,
whose average document length is 3k--8k words.
Our findings suggest that the approach can achieve state-of-the-art performance
and demonstrate better transferability when equipped with a segmentation component.
Our contributions are summarized as follows.
\begin{itemize}[topsep=10pt,itemsep=0pt,leftmargin=*]
\item We investigate a new architecture for extractive long document summarization
that has demonstrated a reasonable degree of transferability from written documents to spoken transcripts.

\item Our model learns effective sentence representations by performing section segmentation and summarization in one fell swoop, enhanced by an optimization-based framework that utilizes the determinantal point process to select salient and diverse summary sentences. 

\item The model achieves state-of-the-art performance on publicly available summarization benchmarks.
Further, we conduct a series of analyses to examine why segmentation aids extractive summarization 
of long documents.
Our code and models are available online: \url{https://github.com/tencent-ailab/Lodoss}
\end{itemize}

\section{Related Work}

There is growing interest in generating concise summaries from long documents.
Most summarizers are enabled by Transformer-based models that can process \emph{long sequences}. 
E.g., Longformer~\cite{Beltagy:2020} replaces Transformer's self-attention mechanism with dilated sliding window attention to reduce computation and memory usage. 
Other methods include content-based and temporal sparse attention~\cite{child2019generating,zaheer2020bigbird,roy-etal-2021-efficient,huang-etal-2021-efficient}
and hierarchical attention that builds representations from words to sentences and eventually to documents~\cite{zhang-etal-2019-hibert,Rohde-etal_2020}.
Our work builds on Longformer to process input documents of substantial length while focusing on probing document structure for summarization.

While abstractive strategies could produce succinct summaries, 
they are prone to hallucinations and factual errors that can mislead the reader~\cite{falke-etal-2019-ranking,kryscinski-etal-2020-evaluating,maynez-etal-2020-faithfulness,pagnoni-etal-2021-understanding}. 
The problem is exacerbated when the inputs are spoken transcripts,
where false starts, repetitions, interjections, ungrammatical sentences are abundant~\cite{Shriberg:1994}.
They may cause errors to propagate through abstractive systems~\cite{shang-etal-2018-unsupervised,li-etal-2019-keep,zhu-etal-2020-hierarchical,koay-etal-2020-domain,koay-etal-2021-sliding,zhong-etal-2021-qmsum,chen-etal-2022-summscreen}. 
Instead, we pursue a more flexible strategy to produce extractive summaries, allowing the reader to grasp the essentials without having to read all materials.

Our work differs from previous extractive methods in its focus on document segmentation,
which holds promise for summarizing lengthy documents.
Important sentences are often located at the beginning or end of documents~\cite{Baxendale:1958,marcu-1998-improving}.
This simple heuristic gives strong results on news summarization~\cite{kedzie-etal-2018-content,chen-bansal-2018-fast,narayan-etal-2018-ranking,mao2021dyle,liu-etal-2022-brio}.
We take one step further, jointly partitioning a document into multiple sections and 
estimating sentence salience given their proximity to section boundaries.
We then explore segmentation of written and spoken documents to understand the model's transferability.

Previous studies rely heavily on lexical cohesion to perform text segmentation~\citep{hearst-1997-text,passonneau-litman-1997-discourse,malioutov-barzilay-2006-minimum}.
Despite their success, establishing coherence in a text goes beyond repeating keywords.
A good writer often use \emph{discourse cues} to create parallel structures, give examples, compare and contrast, or show addition. 
Our method draws inspiration from neural text segmentation models to predict section boundaries~\citep{koshorek-etal-2018-text,arnold-etal-2019-sector,xing-etal-2020-improving,lukasik-etal-2020-text}.
In particular, \citet{koshorek-etal-2018-text} learn sentence representations and label each sentence as ending a segment or not.
\citet{lukasik-etal-2020-text} compare three model architectures based on Transformers and report results on a Wikipedia dataset.
In this paper, we focus on unifying section segmentation and summarization into a single optimization framework, 
exploiting document structure to accurately locate salient content.
In what follows, we describe our approach in greater detail.

\section{Our Approach}
\label{sec:approach}

Let $\mathcal{D} = \{s_1, s_2, ..., s_N\}$ be a document containing $N$ sentences.\footnote{
Modern speech-to-text services provide automatic punctuation.
Transcripts are punctuated using commas, periods, question marks and semicolons. 
It allows us to break down a transcript into a sequence of utterances 
akin to sentences of a written document.
}
Our goal is to create an extractive summary of the document by selecting a subset of $K$ sentences that retains the most important information.
The task of long document summarization is significantly more challenging than other summarization tasks~\cite{daume-iii-marcu-2002-noisy}.
It has a high compression rate, e.g., >85\%,
excluding most sentences from the summary
and suggesting an extractive summarizer must be able to accurately identify summary-worthy sentences.

Fig.~\ref{fig:network} shows a schematic overview of our system named ``\textbf{Lodoss}'' (\textsl{\textbf{Lo}ng} \textsl{\textbf{do}cument} \textsl{\textbf{s}ummarization} \textsl{with} \textsl{\textbf{s}egmentation}).
It learns robust sentence representations by performing both tasks simultaneously.
Further, it introduces a new regularizer drawing on determinantal point processes \cite{cho-etal-2019-multi,DPPAttn:2021} to measure the quality of all summary sentences collectively, ensuring they are informative and have minimum redundancy.

We employ the Longformer~\cite{Beltagy:2020} equipped with dilated window attention to 
produce contextualized token embeddings for an input document.
Windowed attention allows each token to attend only to its local window 
to reduce computation and memory usage.
It has the added benefit of easing section segmentation.
The left and right context of a section break can be captured by the local window,
which reveals any words they have in common and new words seen in this context.
Our Longformer model utilizes a large position embeddings matrix, allowing it to process long documents up to 16K tokens.
We use dilation, changing window size across layers from 32 (bottom) to 512 (top)
to increase its receptive field.

Our summarizer is built on top of Longformer by stacking two layers of inter-sentence Transformers to it.
We append \textsc{[cls]} to the beginning of each sentence and \textsc{[sep]} to the end,
following the convention of \cite{liu-lapata-2019-text}.
This modified sequence of tokens is sent to Longformer for token encoding. 
We obtain the vector of the $i$-th \textsc{[cls]} token as the representation for sentence $s_i$ with rich contextual information.
These vectors are added to sinusoidal position embeddings, 
then passed to two layers of inter-sentence Transformers 
to capture document-level context.
Such global context is especially important for identifying salient sentences,
whereas sinusoidal position embeddings indicate the relative position of sentences.
The output vectors are denoted by $\{\mathbf{h}_{i}\}_{i=1}^N$.

\vspace{0.04in}
\noindent \textbf{\textsl{Summarization and Section Segmentation}}.\\
We address both problems simultaneously in a single framework (Eq.~(\ref{eq:pred_sum}-\ref{eq:pred_seg})). 
Particularly, $y_{\textsl{sum},i} = 1$ indicates the $i$-th sentence is to be included in the summary;
$y_{\textsl{seg},i} = 1$ suggests the sentence starts (or ends) a section.
Both tasks are related at their core.
A section usually starts or concludes with summary-worthy sentences;
predicting section boundaries helps us effectively locate those sentences.
Moreover, the discourse cues for identifying major section boundaries, e.g., ``\emph{so next we need to...},''
are portable across domains and genres.
It allows us to perform a series of ablations to adapt our summarizer from written to spoken documents.
\begin{align*}
& \hat{y}_{\textsl{sum},i} = \sigma ( \mathbf{w}^{\top}_{\textsl{sum}} \mathbf{h}_{i} + b_{\textsl{sum}} )
\numberthis\label{eq:pred_sum}\\
& \hat{y}_{\textsl{seg},i} = \sigma ( \mathbf{w}^{\top}_{\textsl{seg}} \mathbf{h}_{i} + b_{\textsl{seg}} )
\numberthis\label{eq:pred_seg}
\end{align*}

Our base model, ``\textbf{Lodoss}-\emph{base},'' minimizes the per-sentence empirical cross-entropy of the model w.r.t. gold-standard summary labels (Eq.~(\ref{eq:loss_sum})).
It learns to identify salient sentences despite that content salience may vary across datasets.
Further, our joint model, ``\textbf{Lodoss}-\emph{joint},'' optimizes both tasks through multi-task learning: $\mathcal{L}(\Theta) = \mathcal{L}_{\textsl{sum}} + \mathcal{L}_{\textsl{seg}}$.
It adds to the robustness of derived sentence representations, 
because the acquired knowledge for section segmentation is more transferable across domains.
Here, $\hat{y}_{\textsl{sum},i}$ and $\hat{y}_{\textsl{seg},i}$ and are predicted scores for summarization and segmentation; 
$y_{\textsl{sum},i}$ and $y_{\textsl{seg},i}$ are ground-truth sentence labels.
\begin{align*}
& \mathcal{L}_{\textsl{sum}} = - \frac{1}{N} \sum_{i=1}^N \Big( y_{\textsl{sum},i} \log \hat{y}_{\textsl{sum},i} \\   
& \quad\quad\quad\quad + (1-y_{\textsl{sum},i}) \log (1- \hat{y}_{\textsl{sum},i}) \Big)
\numberthis\label{eq:loss_sum}\\
& \mathcal{L}_{\textsl{seg}} = - \frac{1}{N}\sum_{i=1}^N \Big( y_{\textsl{seg},i} \log \hat{y}_{\textsl{seg},i} \\   
& \quad\quad\quad\quad + (1-y_{\textsl{seg},i}) \log (1- \hat{y}_{\textsl{seg},i}) \Big)
\numberthis\label{eq:loss_seg}
\end{align*}

\noindent\textbf{\textsl{A DPP Regularizer.}}\quad
It is especially important to \emph{collectively} measure the quality of a set of extracted sentences, instead of handling sentences individually.
We introduce a new regularizer leveraging the determinantal point processes~\cite{Kulesza:2012,Zhang:2016:DPP,DPPAttn:2021}
to encourage a set of salient and diverse sentences to be selected for the summary.
With the DPP regularizer, a ground-truth summary $Y^{'}$ is expected to achieve the highest probability score compared to alternatives.
It provides a summary-level training objective that complements the learning signals of our \textbf{Lodoss}-\emph{joint} summarizer.

DPP defines a probabilistic measure for scoring a subset of sentences.
Let $\mathcal{Y} = \{1,2,...,\textsf{N}\}$ be the ground set containing \textsf{N} sentences. 
The probability of a subset $Y \subseteq \mathcal{Y}$, corresponding to an extractive summary, is given by Eq.~(\ref{eq:p_y}),
where $\mbox{det}(\cdot)$ is the determinant of a matrix;
$L \in \mathbb{R}^{\textsf{N} \times \textsf{N}}$ is a positive semi-definite matrix; 
$L_{ij}$ indicates the similarity between sentences $i$ and $j$;
$L_Y$ is a principal minor of $L$ indexed by elements in $Y$;
$I$ is an identity matrix of the same dimension as $L$.
\begin{align*}
&\mathcal{P}(Y) = \frac{\mbox{det}(L_Y)}{\mbox{det}(L + I)}
\numberthis\label{eq:p_y}
\end{align*}

We make use of the quality-diversity decomposition for constructing $L$:
$L = \mbox{diag}(\mathbf{q}) \cdot \mathbf{S} \cdot \mbox{diag}(\mathbf{q})$,
where $\mathbf{q} \in \mathbb{R}^\textsf{N}$ represents the quality of sentences;
$\mathbf{S} \in \mathbb{R}^{\textsf{N} \times \textsf{N}}$ captures the similarity of sentence pairs.
In our model, the sentence quality score $q_i$ is given by $\hat{y}_{\textsl{sum},i}$ (Eq.~(\ref{eq:pred_sum})),
indicating its importance to the summary.
The sentence similarity score is defined by: 
$S_{i,j} = \cos(\mathbf{h}_i, \mathbf{h}_j) = \frac{\mathbf{h}_{i}\mathbf{h}_{j}}{\norm{\mathbf{h}_i} \times \norm{\mathbf{h}_j}}$.
We employ batch matrix multiplication (BMM) to efficiently perform batch matrix-matrix products.

DPP rewards a summary if it contains a subset of important and diverse sentences.
A summary containing two sentences $i$ and $j$ 
has a high probability score $\mathcal{P}(Y=\{i,j\})$ if the sentences are of high quality
and dissimilar from each other.
Conversely, if two identical sentences are included in the summary,
the determinant $\mbox{det}(L_{Y})$ is zero.
Modeling pairwise repulsiveness
helps increase the diversity of the selected sentences
and eliminate redundancy.
As illustrated in Eq.~(\ref{eq:loss_dpp}),
our DPP regularizer is defined as the negative log-probability of the ground-truth extractive summary $Y^{'}$.
It has the practical effect of promoting selection of the ground-truth summary 
while down-weighting alternatives. 
\begin{align*}
& \mathcal{L}_{\textsl{DPP}} = -\log \frac{\mbox{det}(L_{Y^{'}})}{\mbox{det}(L + I)} 
\numberthis\label{eq:loss_dpp}
\end{align*}

Our final model, ``\textbf{Lodoss}-\emph{full},'' is shown in Figure~\ref{fig:network}. 
It adds the DPP regularizer to the joint model (Eq.~(\ref{eq:loss_all}));
$\beta$ is a coefficient that balances sentence-level cross-entropy losses and summary-level DPP regularization.
$\Theta$ are all of our model parameters.
\begin{align*}
&\mathcal{L}(\Theta) = (\mathcal{L}_{\textsl{sum}} + \mathcal{L}_{\textsl{seg}}) + \beta \mathcal{L}_{\textsl{DPP}}
\numberthis\label{eq:loss_all}
\end{align*}

\section{Experiments}
\label{sec:expr}

In this section, we detail our experimental settings for long document extractive summarization.
Our datasets include collections of scientific articles and lecture transcripts, 
their associated summaries and section boundaries. 
We contrast our approach with strong summarization baseline systems and report results on three standard benchmarks.
Model ablations and human assessment of summary quality are presented in \S\ref{sec:ablation}.

\subsection{Datasets}

For written documents, we choose to experiment with scientific articles~\cite{cohan-etal-2018-discourse} as they follow a logical document structure.
They come with human summaries and sections, delimited by top-level headings.
The scientific articles are gathered from two open-access repositories: \href{arXiv.org}{arXiv.org} and \href{PubMed.com}{PubMed.com}.
Particularly, arXiv consists of papers in the fields of mathematics, physics, astronomy, electrical engineering, computer science, and more.
PubMed contains research articles and their abstracts on life sciences and biomedical topics.
These datasets contain 148K and 216K instances, respectively (Table~\ref{tab:dataset_stat}). 
Their source documents are an order of magnitude longer than standard news articles~\cite{grusky-etal-2018-newsroom}.

For transcript summarization, we utilize lectures gathered from \href{VideoLectures.NET}{VideoLectures.NET}~\cite{lv2021vt}.
These lectures have been automatically transcribed using Microsoft's Speech-to-Text API. 
Further, the transcripts are time aligned with lecture slides.
All utterances aligned to a single slide are grouped into a cluster, they form a transcript \emph{section}.
Text extracted from slides are used as ground-truth summaries. 
The dataset contains a total of 9,616 videos.
Each video contains about 33 slides.
It helps us lay the groundwork for unifying summarization and segmentation on spoken documents, 
using lecture slides to provide weak supervision for both tasks.\footnote{
We investigated other transcript datasets~\cite{Carletta:2006} and found they do not contain enough training instances.
}
We show data statistics in Table~\ref{tab:dataset_stat},
including number of sentences/words per document and summary,
reported for train, validation and test sets.

\begin{table}[t]
\setlength{\tabcolsep}{4pt}
\renewcommand{\arraystretch}{1.1}
\centering
\begin{footnotesize} 
\begin{tabular}{|ll|rrr|rr|}
\hline
\multicolumn{2}{|c|}{} & \multicolumn{3}{c|}{\textbf{\textsl{Document}}} & \multicolumn{2}{c|}{\textbf{\textsl{Summary}}} \\
 & \#Insts & \#Wds & \#Tkns & \#Snts& \#Wds & \#Snts \\
\hline
\hline
\rowcolor{gray!10}
& & \multicolumn{3}{c|}{\textsl{\textbf{PubMed}}} & & \\
Train & 134,915 & 3,044 & 3,865 & 86.3 & 202 & 6.8 \\
Val & 6,633 & 3,112 & 3,982 & 87.9 & 203 & 6.8 \\
Test & 6,658 & 3,093 & 3,914 & 87.5 & 205 & 6.9 \\
\hline
\hline
\rowcolor{gray!10}
& & \multicolumn{3}{c|}{\textsl{\textbf{arXiv}}} & & \\
Train & 203,037 & 6,038 & 8,583 & 206.4 & 280 & 9.9 \\
Val & 6,436 & 5,894 & 8,152 & 204.2 & 162 & 5.6 \\
Test & 6,440 & 5,906 & 8,132 & 205.7 & 163 & 5.7 \\
\hline
\hline
\rowcolor{gray!10}
& & \multicolumn{3}{c|}{\textsl{\textbf{VideoLec}}} & & \\
Train & 7,692 & 4,192 & 4,901 & 291.9 & 456 & 24.5 \\
Val & 962 & 4,222 & 4,931 & 294.1 & 466 & 24.7 \\
Test & 962 & 4,387 & 5,131 & 306.2 & 479 & 25.4 \\
\hline
\end{tabular}
\end{footnotesize}
\caption{Statistics of our datasets.
\#Wds, \#Tkns and \#Snts are average number of words, tokens and sentences, respectively.
We report these for training, validation and test sets.
Tokenization was performed using BPE~\cite{sennrich-etal-2016-neural}.}
\label{tab:dataset_stat}
\end{table}

\vspace{0.04in}
\noindent\textbf{\textsl{Ground-Truth Labels.}}\quad
A label $y_{\textsl{sum},i}$ is assigned to each sentence of the document:
1 indicates the sentence belongs to the \textsc{Oracle} summary, 0 otherwise.
An \textsc{Oracle} is created by adding one sentence at a time incrementally to the summary, so that it improves the average of ROUGE-1 and -2 F-scores~\cite{kedzie-etal-2018-content}.
\textsc{Oracle} summaries give the ceiling performance of an extractive summarizer. 
\textsc{Oracle} summaries for scientific papers are created by us; those for lecture transcripts are provided by \citet{lv2021vt} generated by aligning transcript utterances with lecture slides. 

Scientific papers come with sections:
we specify $y_{\textsl{seg},i}=1$ if it is the first (or last) sentence of a section, 0 otherwise.
Both the first and last sentences of a section could contain discourse connectives indicating a topic shift.
Clear document structure depends on establishing where a section ends and the next one begins.
For lectures, all transcript utterances are time aligned with lecture slides, 
creating mini-sections. 
We explore alternative definitions of transcript sections in \S\ref{sec:results}.

\vspace{0.04in}
\noindent\textbf{\textsl{System Predictions.}}\quad
At inference time, our system extracts a fixed number of sentences ($K$) from an input document.
These sentences have the highest probability scores according to Eq.~(\ref{eq:pred_sum}).\footnote{
Our training objective focuses on learning robust sentence representations (Eq.~(\ref{eq:loss_all})).
We choose to extract sentences based on such representations over a DPP inference algorithm. 
The latter is reported to give fewer summary sentences, yielding high precision but low recall scores~\cite{Zhang:2016:DPP}.
}
$K$ is chosen to be close to the average number of sentences per reference summary.
We set $K$=7 and 5 for the PubMed and arXiv datasets, respectively, following the convention of~\citet{xiao-carenini-2019-extractive}.
We use $K$=3 for lectures~\cite{lv2021vt}.
Section predictions are given by Eq.~(\ref{eq:pred_seg}).

\subsection{Experimental Settings}
\label{sec:settings}

Our implementation uses HuggingFace~\cite{wolf-etal-2020-transformers}, PyTorch~\cite{PyTorch:NEURIPS2019_9015} and PyTorch Lightning~\cite{falcon2019pytorch}.
We use the Adam optimizer. Its initial learning rate is set to be $3e^{-5}$. 
The learning rate is linearly warmed up for 10\% of the total training steps.
The training was performed on 8 NVIDIA Tesla P40 GPUs.
The models were trained on each dataset for 20 epochs,
using a batch size of 8 with gradient accumulation every 4 steps.
We run hyperparameter search trials on the validation set, with $\beta \in \{1, \underline{0.1}, 0.01, 0.001\}$.
We adopt half-precision (FP16) to speed up training for all models,
with the exception of the full model, 
where full-precision (FP32) is used to ensure a stable performance of eigenvalue decomposition required by the DPP regularizer. 
The best results are with 16K tokens.
We use 4K input for all ablations to save computation unless otherwise noted.

\begin{table}[t]
\setlength{\tabcolsep}{3pt}
\renewcommand{\arraystretch}{1.15}
\centering
\begin{footnotesize}
\begin{tabular}{|l|ccc|ccc|}
\hline
& \multicolumn{3}{c|}{\textbf{\textsl{PubMed}}} & \multicolumn{3}{c|}{\textbf{\textsl{arXiv}}}\\
\textbf{System} & \textbf{R-1} & \textbf{R-2} & \textbf{R-L} & \textbf{R-1} & \textbf{R-2} & \textbf{R-L} \\
\hline
\hline
\rowcolor{gray!10}
\multicolumn{7}{|c|}{\textbf{\textsl{Abstractive Systems}}} \\
Discourse & 38.93 & 15.37 & 35.21   &  35.80 & 11.05 & 31.80 \\
TLM-I+E & 42.13 & 16.27 & 39.21     & 41.62 & 14.69 & 38.03 \\
BigBird-\emph{base} & 43.70 & 19.32 & 39.99     & 41.22 & 16.43 & 36.96 \\
BigBird-\emph{large} & 46.32 & 20.65 & 42.33    & 46.63 & 19.02 & 41.77 \\
LED-4K & -- & -- & -- & 44.40 & 17.94 & 39.76 \\
LED-16K & -- & -- & -- & 46.63 & {19.62} & 41.83 \\
HAT & 48.25 & 21.35 & 36.69    & 46.74 & 19.19 & {42.20} \\
\hline
\hline
\rowcolor{gray!10}
\multicolumn{7}{|c|}{\textbf{\textsl{Extractive Systems}}} \\
\textsc{Oracle} & 61.49 & 34.70 & 55.92   & 59.41 & 30.05 & 52.34 \\
\textsc{Lead}-10 & 37.45 & 14.19 & 34.07   & 35.52 & 10.33 & 31.44 \\
SumBasic & 37.15 & 11.36 & 33.43    & 29.47 & 6.95 & 26.30 \\
LexRank & 39.19 & 13.89 & 34.59    & 33.85 & 10.73 & 28.99 \\
ExtSum-LG & 44.85 & 19.70 & 31.43    & 43.62 & 17.36 & 29.14 \\
... + RdLoss & 45.39 & 20.37 & 40.99    & 44.01 & 17.79 & 39.09 \\
Sent-PTR & 45.01 & 19.91 & 41.16    & 42.32 & 15.63 & 38.06  \\ 
\hline
\hline
\rowcolor{gray!10}
\multicolumn{7}{|c|}{\textbf{\textsl{Our System (Extractive)}}} \\
\textbf{Lodoss}-\emph{base} & 48.10 & 22.53 & 43.51     & 47.64 & 19.73 & 41.71 \\ 
\textbf{Lodoss}-\emph{joint} & 48.83 & 23.13 & 44.23    & 47.97 & 20.13 & 42.03 \\ 
\textbf{Lodoss}-\emph{full} & \textbf{48.93} & \textbf{23.51} & \textbf{44.40}    & \textbf{48.20} & \textbf{20.50} & \textbf{42.28}\\ 
\textbf{Lodoss}-\emph{full}-LG & \textbf{49.38} & \textbf{23.89} & \textbf{44.84}    & \textbf{48.45} & \textbf{20.72} & \textbf{42.55}\\ 
\hline
\end{tabular}
\end{footnotesize}
\vspace{-0.05in}
\caption{ROUGE results on the PubMed and arXiv datasets. 
}
\label{tab:results_rouge_arxiv_pubmed_16k}
\end{table}

\subsection{Summarization Results}
\label{sec:results}

\noindent\textbf{\textsl{Baseline Systems.}}\quad
We compare our system with strong summarization baselines.
\textsl{SumBasic}~\cite{Vanderwende:2007} is an extractive approach that adds sentences to the summary if they contain frequently occurring words.
\textsl{LexRank}~\cite{Erkan:2004} measures sentence salience based on eigenvector centrality.
\textsl{ExtSum-LG}~\cite{xiao-carenini-2019-extractive,xiao-carenini-2020-systematically} leverages local and global context to extract salient sentences.
\textsl{+RdLoss} further adds a redundancy loss term to the learning objective to help the model eliminate redundancy in long document summarization.
\textsl{Sent-PTR}~\cite{pilault-etal-2020-extractive} uses a hierarchical seq2seq sentence pointer model for sentence extraction.

Our abstractive baselines include the following:
\textsl{Discourse}~\cite{cohan-etal-2018-discourse} utilizes a hierarchical encoder to model the document structure and an attentive decoder to generate the summary.
\textsl{TLM-I+E} ~\cite{pilault-etal-2020-extractive} generates a paper abstract using the Transformer language model, where the introduction section and extracted sentences are provided as context.
\textsl{BigBird}~\cite{zaheer2020bigbird} and \textit{LED}~\cite{Beltagy:2020} use sparse attention and windowed attention to process long input sequences.
\textsl{HAT}~\cite{Rohde-etal_2020} adds hierarchical attention layers to an encoder-decoder model to summarize long documents. 

\vspace{0.05in}
\noindent\textbf{\textsl{Results on Scientific Papers.}}\quad
We compare three of our model variants, listed below. 
Standard evaluation metrics (ROUGE; Lin, 2004)\nocite{lin-2004-rouge},
including R-1, R-2 and R-L, are used to measure the quality of system summaries.
It allows our model to be directly compared to previous approaches.
More ablations and human assessment are provided in \S\ref{sec:ablation}.
\begin{itemize}[topsep=3pt,itemsep=0pt,leftmargin=*]
\item \textbf{Lodoss}-\emph{base}, using $\mathcal{L}_{\textsl{sum}}$
\item \textbf{Lodoss}-\emph{joint}, using $\mathcal{L}_{\textsl{sum}}+\mathcal{L}_{\textsl{seg}}$
\item \textbf{Lodoss}-\emph{full}, using $(\mathcal{L}_{\textsl{sum}}+\mathcal{L}_{\textsl{seg}})+\beta\mathcal{L}_{\textsl{DPP}}$
\end{itemize}

\begin{table}
\setlength{\tabcolsep}{3.2pt}
\renewcommand{\arraystretch}{1.1}
\centering
\scriptsize
\begin{small}
\begin{tabular}{|ll|ccc|ccc|}
\hline
& \textbf{System} & \textbf{P} & \textbf{R} & \textbf{F} & \textbf{R-1} & \textbf{R-2} & \textbf{Avg(R)} \\
\hline
\hline
\multirow{4}{*}{\rotatebox[origin=c]{90}{None}} & LexRank & 17.38 & 3.66 & 5.07 & 17.07 & 5.78 & 10.02\\
& TextRank & 21.26 & 4.38 & 6.10 & 20.50 & 6.68 & 11.85\\
& Lo-jnt-sgl & 43.94 & 7.83 & 12.09 & 23.86 & 15.11 & 20.60\\
& Lo-fll-sgl & 47.87 & 8.48 & 13.11 & 24.12 & 15.85 & 21.04\\
\hline
\hline
\multirow{4}{*}{\rotatebox[origin=c]{90}{arXiv}} & Lo-jnt-sgl & 46.44 & 8.12 & 12.63 & 24.46 & 15.93 & 21.29\\
& Lo-fll-sgl & 47.18 & 8.39 & 12.97 & 24.38 & 16.04 & 21.28\\
& \cellcolor{teal!18}Lo-jnt-grp & \cellcolor{teal!18}49.31 & \cellcolor{teal!18}8.80 & \cellcolor{teal!18}{13.59} & \cellcolor{teal!18}25.01 & \cellcolor{teal!18}16.93 & \cellcolor{teal!18}{22.02}\\ 
& \cellcolor{teal!18}Lo-fll-grp & \cellcolor{teal!18}48.00 & \cellcolor{teal!18}8.30 & \cellcolor{teal!18}{12.95} & \cellcolor{teal!18}24.34 & \cellcolor{teal!18}16.18 & \cellcolor{teal!18}{21.31}\\ 
\hline
\hline
\multirow{4}{*}{\rotatebox[origin=c]{90}{PubMed}} & Lo-jnt-sgl & 48.11 & 8.44 & 13.02 & 24.76 & 16.27 & 21.57\\
& Lo-fll-sgl & 47.69 & 8.52 & 13.08 & 24.61 & 16.24 & 21.49\\
& \cellcolor{teal!18}Lo-jnt-grp & \cellcolor{teal!18}51.00 & \cellcolor{teal!18}9.17 & \cellcolor{teal!18}{14.10} & \cellcolor{teal!18}24.89 & \cellcolor{teal!18}16.88 & \cellcolor{teal!18}{21.90}\\
& \cellcolor{teal!18}Lo-fll-grp & \cellcolor{teal!18}49.29 & \cellcolor{teal!18}8.97 & \cellcolor{teal!18}{13.69} & \cellcolor{teal!18}24.72 & \cellcolor{teal!18}16.63 & \cellcolor{teal!18}{21.72}\\
\hline
\end{tabular}
\end{small}
\vspace{-0.05in}
\caption{Results on lecture transcripts.
The metrics reported are Precision, recall, F-scores, and Rouge scores. 
Our model can be trained from scratch, or pretrained on either arXiv or PubMed.
We explore alternative definitions of \emph{sections}:
all utterances aligned to a single slide is a section (`sgl') vs.
using six major sections per transcript (`grp').
}
\label{tab:rouge_vtssum}
\end{table}

\begin{table*}[!t]
\setlength{\tabcolsep}{3.1pt}
\renewcommand{\arraystretch}{1.15}
\centering
\begin{footnotesize}
\begin{tabular}{|lr||lll|lll|l||lll|lll|l|}
\hline
& & \multicolumn{6}{c|}{\textbf{\textsl{PubMed}}} & \multirow{3}{3em}{\textbf{\#Wds}} & \multicolumn{6}{c|}{\textbf{\textsl{arXiv}}} & \multirow{3}{3em}{\textbf{\#Wds}}\\
& & \multicolumn{3}{c|}{\textbf{ROUGE-1}} & \multicolumn{3}{c|}{\textbf{ROUGE-2}} &  & \multicolumn{3}{c|}{\textbf{ROUGE-1}} & \multicolumn{3}{c|}{\textbf{ROUGE-2}} & \\
& \textbf{Lodoss} & \textbf{P}(\%) & \textbf{R}(\%) & \textbf{F}(\%) & \textbf{P}(\%) & \textbf{R}(\%) & \textbf{F}(\%) &  &  \textbf{P}(\%) & \textbf{R}(\%) & \textbf{F}(\%) & \textbf{P}(\%) & \textbf{R}(\%) & \textbf{F}(\%) & \\
\hline
\hline
\multirow{3}{*}{\rotatebox[origin=c]{90}{5-Sent}} & \emph{base} & 50.25 & 49.56 & 47.75 & 23.28 & 22.19 & 21.73 & 204.3 & 42.10 & 55.13 & 46.04 & 16.71 & 21.84 & 18.24 & 216.5\\
& \emph{joint} & 50.75 & 49.31 & 47.85\textcolor{red}{$\uparrow$} & 23.72 & 22.24 & 21.96\textcolor{red}{$\uparrow$} & 202.3\textcolor{blue}{$\downarrow$} & 43.56 & 54.21 & 46.50\textcolor{red}{$\uparrow$} & 17.33 & 21.60 & 18.49\textcolor{red}{$\uparrow$} & 204.9\textcolor{blue}{$\downarrow$}\\
& \emph{full} & 51.27 & 49.16 & 48.04\textcolor{red}{$\uparrow$} & 24.01 & 22.26 & 22.11\textcolor{red}{$\uparrow$} & 198.6\textcolor{blue}{$\downarrow$} & 43.37 & 54.61 & 46.59\textcolor{red}{$\uparrow$} & 17.27 & 21.75 & 18.53\textcolor{red}{$\uparrow$} & 207.2\textcolor{blue}{$\downarrow$}\\
\hline
\hline
\multirow{3}{*}{\rotatebox[origin=c]{90}{6-Sent}} & \emph{base} & 47.58 & 53.57 & 48.32 & 22.18 & 24.21 & 22.18 & 235.9 & 38.93 & 58.85 & 45.27 & 15.67 & 23.76 & 18.23 & 250.6\\
& \emph{joint} & 48.06 & 53.26 & 48.43\textcolor{red}{$\uparrow$} & 22.57 & 24.21 & 22.39\textcolor{red}{$\uparrow$} & 233.6\textcolor{blue}{$\downarrow$} & 40.28 & 58.09 & 45.87\textcolor{red}{$\uparrow$} & 16.25 & 23.57 & 18.54\textcolor{red}{$\uparrow$} & 238.5\textcolor{blue}{$\downarrow$}\\
& \emph{full} & 48.57 & 53.17 & 48.65\textcolor{red}{$\uparrow$} & 22.88 & 24.27 & 22.57\textcolor{red}{$\uparrow$} & 229.8\textcolor{blue}{$\downarrow$} & 40.18 & 58.36 & 45.93\textcolor{red}{$\uparrow$} & 16.22 & 23.69 & 18.56\textcolor{red}{$\uparrow$} & 240.1\textcolor{blue}{$\downarrow$}\\
\hline
\hline
\multirow{3}{*}{\rotatebox[origin=c]{90}{7-Sent}} & \emph{base} & 45.19 & 56.64 & 48.21 & 21.22 & 25.82 & 22.33 & 265.8 & 36.23 & 61.76 & 44.17 & 14.80 & 25.41 & 18.08 & 283.4\\
& \emph{joint} & 45.68 & 56.49 & 48.47\textcolor{red}{$\uparrow$} & 21.57 & 25.90 & 22.58\textcolor{red}{$\uparrow$} & 263.1\textcolor{blue}{$\downarrow$} & 37.45 & 61.09 & 44.81\textcolor{red}{$\uparrow$} & 15.33 & 25.22 & 18.39\textcolor{red}{$\uparrow$} & 270.9\textcolor{blue}{$\downarrow$}\\
& \emph{full} & 46.06 & 56.40 & 48.60\textcolor{red}{$\uparrow$} & 21.84 & 25.95 & 22.73\textcolor{red}{$\uparrow$} & 260.4\textcolor{blue}{$\downarrow$} & 37.32 & 61.39 & 44.83\textcolor{red}{$\uparrow$} & 15.28 & 25.37 & 18.40\textcolor{red}{$\uparrow$} & 273.3\textcolor{blue}{$\downarrow$}\\
\hline
\end{tabular}
\end{footnotesize}
\vspace{-0.05in}
\caption{We vary the length of output summaries to contain 5-7 sentences and report summarization results on PubMed and arXiv. 
Our model \textbf{Lodoss}-\emph{full} consistently outperforms other variants across all lengths and evaluation metrics.
}
\label{tab:results_rouge_PRF1}
\end{table*}

Results on PubMed and arXiv datasets are shown in Table~\ref{tab:results_rouge_arxiv_pubmed_16k}.
Our models strongly outperform both extractive and abstractive baselines, 
suggesting the effectiveness of unifying section segmentation with summarization.
The \textsc{Lead} baseline, however, does not perform as well on long documents as it does on news articles.
It is interesting to note that 
our models are trained with indirect signals,
i.e., binary sentence labels derived from reference summaries, 
and they remain quite effective at capturing salient content on long documents.

We conduct significance tests using the approximate randomization method~\cite{riezler-maxwell-2005-pitfalls, significance_test}.
With a confidence level of 99\%, 
all of our Lodoss models are significantly better than BigBird-base and LED-4K.
The differences between our model variants are also significant: 
between \textbf{Lodoss}-\emph{base} and \textbf{Lodoss}-\emph{joint}, between \textbf{Lodoss}-\emph{joint} and \textbf{Lodoss}-\emph{full}. 
Our results indicate that incorporating section segmentation and a summary-level DPP regularizer can help the model better locate salient sentences.
Moreover, the large encoder (`-LG') results in improvements on both datasets.

\vspace{0.05in}
\noindent\textbf{\textsl{Results on Lecture Transcripts.}}\quad
We could train our model from scratch using lecture transcripts, or pretrain the model on either arXiv or PubMed, then fine-tune it on transcripts.
Results are shown in Table~\ref{tab:rouge_vtssum}.
The metrics reported are precision, recall, F-scores and Rouge scores.~\footnote{
Ground-truth abstractive summaries are unavailable for this dataset.
We use sentences labeled as summaries to compute Rouge scores.
We could not directly compare our results to those of~\cite{lv2021vt} due to different settings used. 
} 
We observe that models pretrained on written documents perform substantially better 
compared to training a model from scratch, and PubMed outperforms arXiv consistently except \text{Lo-full-grp}.
It suggests that knowledge gained from summarizing written documents could be transferred to summarization of spoken transcripts. 
This is especially the case for our joint model (Lo-joint-$\star$),
where the models is equipped with the ability to recognize section boundaries.
The Lo-joint-$\star$ model consistently outperforms the model trained from scratch regardless of different segmentation labels.
Note that F-scores are not necessarily aligned with the Rouge scores as the system can predict sentences with similar context that are not labeled as summaries.

We explore alternative definitions of a \emph{transcript section}:
all utterances aligned to a single slide is considered a section vs.
using six major sections per transcript.
The former leads to about 33 sections per transcript. 
The latter is achieved by finding 6 slides that are discussed the most,
and using the first utterance of each slide as the start of a new section. 
Because scientific papers on PubMed and arXiv contain 6.06 and 5.68 sections averagely,
this definition allows our model to be pretrained and fine-tuned 
under similar conditions.
We find that using six major sections per transcript improves summarization performance.

\begin{figure}[t]	
\centering
\includegraphics[width=2.8in]{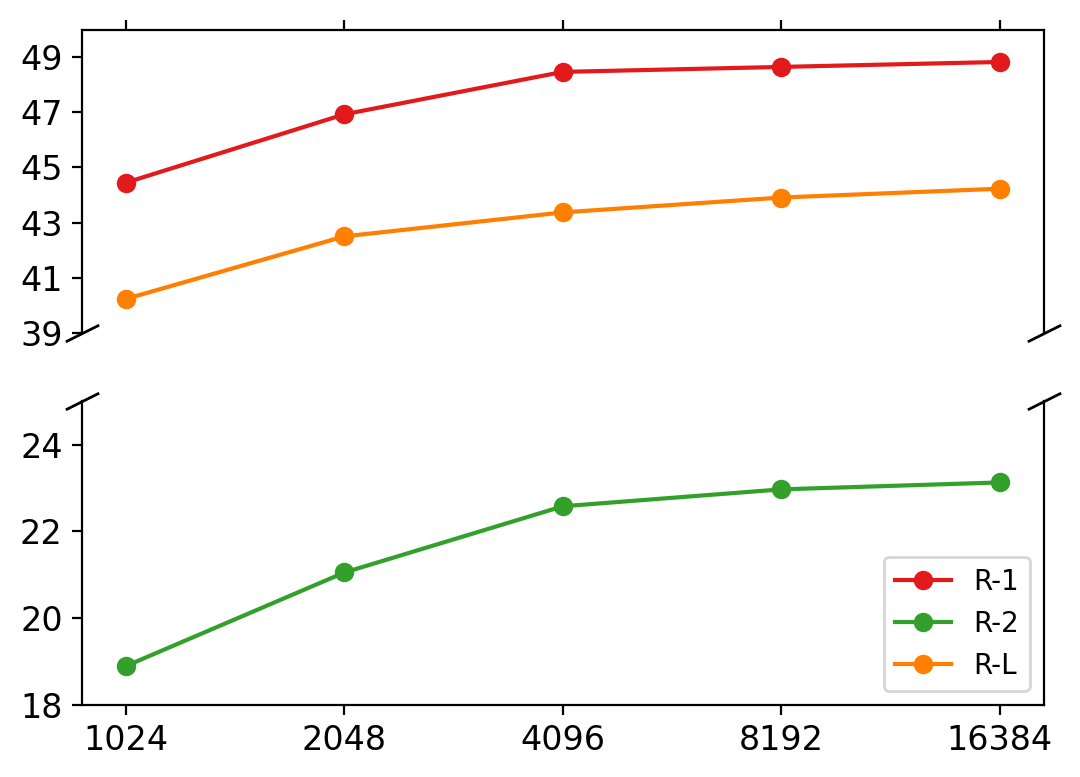}
\caption{
Effect of varying source sequence lengths (PubMed);
x-axis shows the source sequence length measured by number of tokens;
y-axis shows the ROUGE scores.
}
\label{fig:rouge_by_seqlen}
\end{figure}

\section{Ablations and Analyses}
\label{sec:ablation}

\noindent\textbf{\textsl{Effect of Summary Length.}}\quad\,
We vary the length of output summaries to contain 5-7 sentences and report summarization results on PubMed and arXiv (Table~\ref{tab:results_rouge_PRF1}). 
Our model \textbf{Lodoss}-\emph{full} consistently outperforms other variants across all lengths and evaluation metrics.
The highest scores are obtained for PubMed with 7 output sentences, whereas
5 sentences work best for arXiv, as it gives a good tradeoff between recall and precision.

\vspace{0.05in}
\noindent\textbf{\textsl{Effect of Source Sequence Length.}}\quad
We observe that our model performs increasingly better 
when longer source sequences are used (Figure~\ref{fig:rouge_by_seqlen}).
This is expected, as importance information will be left out 
if source sequences need to be truncated to a certain length.
For example, using 4K tokens, we have to truncate about 50\% of arXiv inputs.

\begin{figure}[t]	
\centering
\includegraphics[width=2.5in]{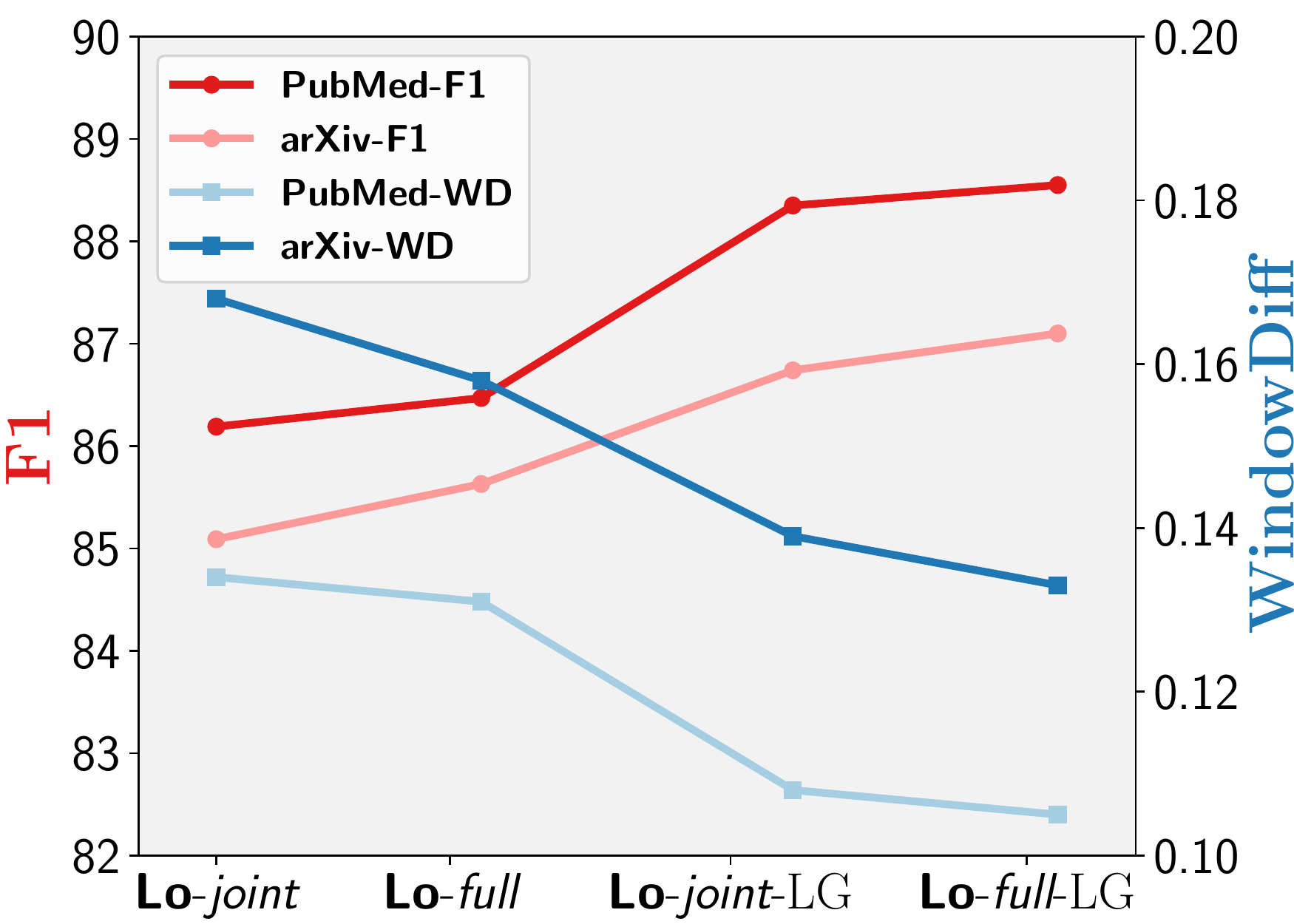}
\caption{Section segmentation results evaluated by F1 (higher is better) and WinDiff (lower is better). Results are reported for PubMed and arXiv. 
Best performance is achieved with our \textbf{Lodoss}-\emph{full} model.
`-LG' means a Longformer-large model is used to encode the input document.
}
\label{fig:seg_results}
\end{figure}

\vspace{0.05in}
\noindent\textbf{\textsl{Model's Performance on Section Segmentation.}}\\
Figure~\ref{fig:seg_results} shows segmentation results for PubMed and arXiv.
Our goal is to predict the first sentence of a section.
F1 and WindowDiff scores \cite{pevzner-hearst-2002-critique} are reported.
Particularly, WindowDiff is a lenient measure for segmentation results.
It uses a sliding window to scan through the input document.
At each step, it examines whether the predicted boundaries are correct within the local window.
We observe that both our full model and large pretrained models help the system to better predict section boundaries.
Predicting the first sentence of a section is easier than predicting the last sentence (Table~\ref{tab:seg_rouge}).
This gives 4\% and 6\% gain, respectively, on PubMed and arXiv.

\begin{table}[!t]
\setlength{\tabcolsep}{4.5pt}
\renewcommand{\arraystretch}{1.1}
\centering
\footnotesize
\begin{footnotesize}
\begin{tabular}{|l|l|cc|cc|}
\hline
\multirow{2}{3em}{\textbf{Model}}& \multirow{2}{1.8em}{\textbf{Pos}} & \multicolumn{2}{c|}{\textbf{PubMed}} & \multicolumn{2}{c|}{\textbf{arXiv}}\\
 &  & \textbf{F1} & \textbf{Avg-R} & \textbf{F1} & \textbf{Avg-R}  \\
\hline
\hline
$\textbf{Lodoss}$-$\emph{joint}$ & \multirow{3}{1.0em}{1st} & 86.19 & 38.73 & 85.09 & 36.71 \\
$\textbf{Lodoss}$-$\emph{full}$ & & 86.47 & 38.95 & 85.63 & 36.99 \\
$\textbf{Lodoss}$-$\emph{full}$-$\texttt{LG}$ & & 88.74 & 39.37 & 87.10 & 37.24 \\
\hline
\hline
$\textbf{Lodoss}$-$\emph{joint}$ & \multirow{3}{1.0em}{Last} & 84.93 & 38.42 & 77.07 & 36.69 \\
$\textbf{Lodoss}$-$\emph{full}$ &  & 85.26 & 38.92 & 78.41 & 36.95 \\
$\textbf{Lodoss}$-$\emph{full}$-$\texttt{LG}$ & & 87.19 & 39.36 & 81.71 & 37.25 \\
\hline
\end{tabular}
\end{footnotesize}
\caption{Segmentation (F1) and Summarization (ROUGE) results using different segmentation labels.}
\label{tab:seg_rouge}
\vspace{-0.1in}
\end{table}

\begin{figure}[!t]	
\centering
\includegraphics[width=3in]{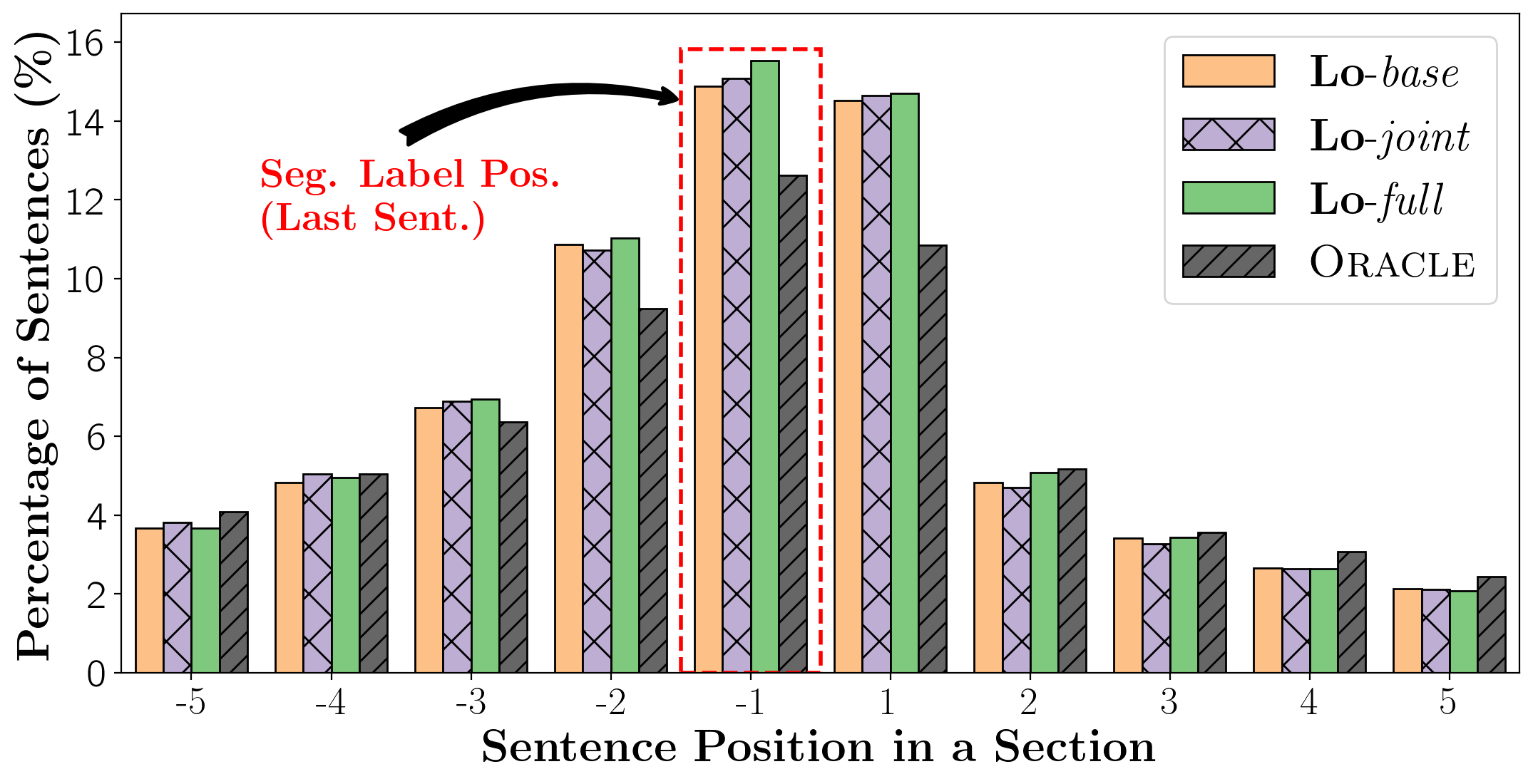}    
\vspace{-0.25in}
\caption{How often summary sentences are found near section boundaries (PubMed). 
``1'' indicates a summary sentence is the first sentence of a section,
whereas ``-1'' indicates it is the last sentence of a section.
Both the first and last sentences of a section are likely to be
selected for inclusion in the summary.
}
\label{fig:rela_sent_pos_pubmed}
\end{figure}

\begin{figure}[!t]	
	\centering
	\includegraphics[width=3in]{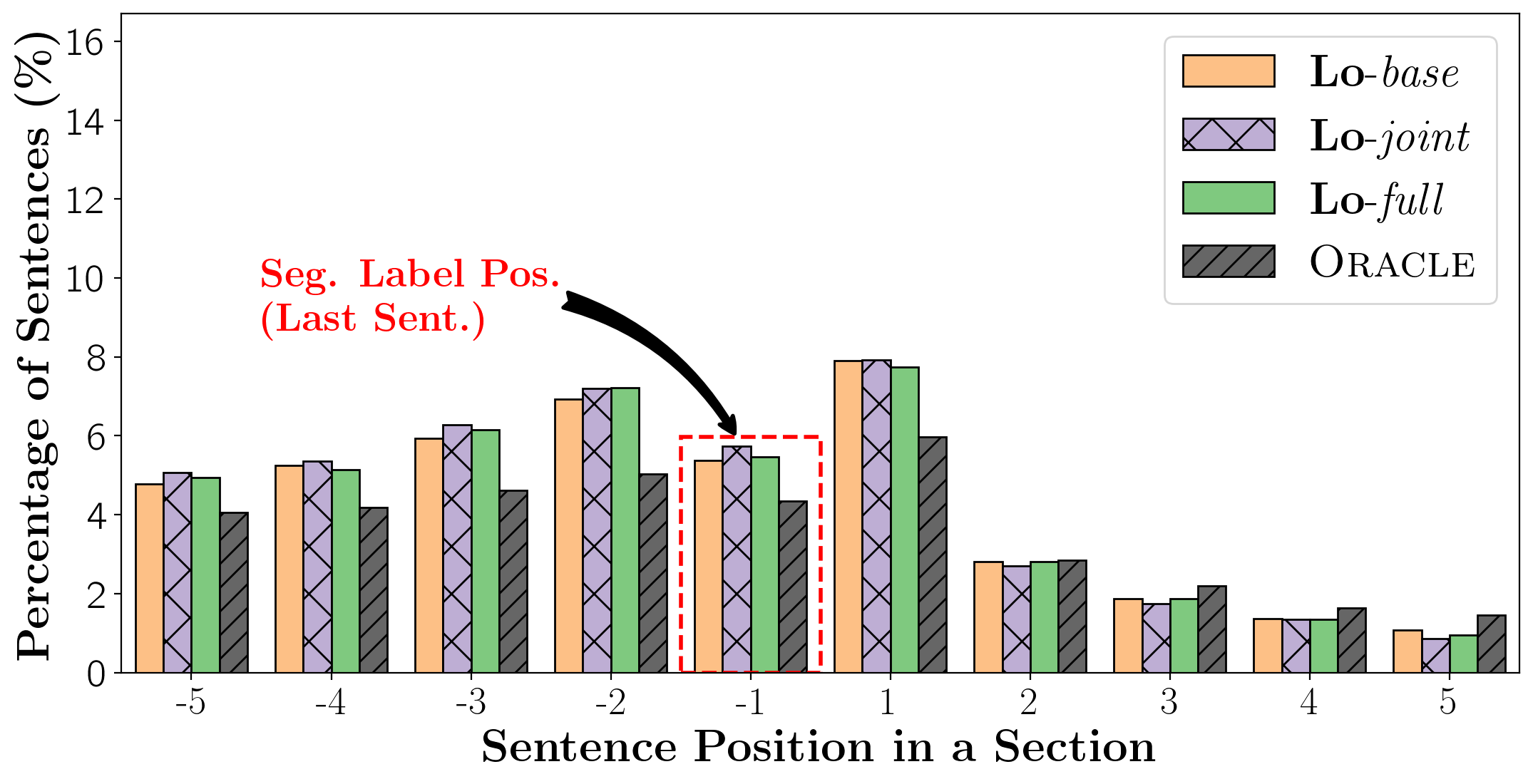}  
	\vspace{-0.25in}
	\caption{How often summary sentences are found near section boundaries (arXiv).}
	\label{fig:rela_sent_pos_arxiv}
\end{figure}

\vspace{0.05in}
\noindent\textbf{\textsl{Effect of Our DPP Regularizer.}}\quad\,
Table~\ref{tab:results_rouge_PRF1} shows the average number of words per summary,
where summaries are produced by different model variants.
We find that summaries produced by \textbf{Lodoss}-\emph{full}
tend to be shorter compared to other summaries,
and \textbf{Lodoss}-\emph{full} remains the best performing model.
It suggests that the DPP regularizer favors a diverse set of sentences to be included in the summary. 
The selected sentences are not necessarily long as they may contain redundant content.

\vspace{0.05in}
\noindent\textbf{\textsl{Why Section Segmentation is Necessary.}}\quad
We investigate how often summary sentences are found near section boundaries. 
Results are shown in Figure~\ref{fig:rela_sent_pos_pubmed} and ~\ref{fig:rela_sent_pos_arxiv}, respectively for PubMed and arXiv.
``1'' indicates a summary sentence is the first sentence of a section,
whereas ``-1'' indicates it is the last sentence of a section.
Overall, both the first and last sentences of a section are likely to be
selected for inclusion in the summary.
The effect is stronger for PubMed and less so for arXiv. 
We conjecture that because arXiv papers are twice as long as PubMed papers,
summary sentences may not always occur near section boundaries. 
In both cases, our models are able to leverage this characteristic to simultaneously identify summary sentences and perform section segmentation.

\begin{table}[!t]
\setlength{\tabcolsep}{3.8pt}
\renewcommand{\arraystretch}{1.1}
\centering
\footnotesize
\begin{small}
\begin{tabular}{|ll|c|c|c|c|}
\hline
& \textbf{Model} & \textbf{Avg.} & \textbf{4/5} & \textbf{3} & \textbf{1/2} \\
\hline
\hline
\multirow{2}{*}{{Info$\uparrow$}} & \textbf{Lodoss}-\emph{joint}& 2.89 & 10.05\% & 21.84\% & 68.11\%\\
& \textbf{Lodoss}-\emph{full}& \textbf{3.12} & \textbf{14.70}\% & 21.49\% & \textbf{63.81}\%\\
\hline
\hline
\multirow{2}{*}{{Div$\downarrow$}} & \textbf{Lodoss}-\emph{joint}& 2.14  & 5.50\% & 13.16\% & 81.34\%\\
& \textbf{Lodoss}-\emph{full}& \textbf{2.03} & \textbf{3.40}\% & 14.06\% & \textbf{82.54}\%\\
\hline
\end{tabular}
\end{small}
\vspace{-0.05in}
\caption{Human evaluation results for informativeness (higher is better) and diversity (lower is better).}
\label{tab:human_eval}
\end{table}

\vspace{0.05in}
\noindent\textbf{\textsl{Human Assessment of System Summaries.}}\quad
We focus on evaluating informativeness and diversity of summary sentences.
Other criteria are not considered because extractive summaries 
can be highlighted on their source materials, allowing them to be understood in context.
Our evaluation focuses on the \textbf{Lodoss}-\emph{joint} and \textbf{Lodoss}-\emph{full} models. 
The informativeness and diversity metrics are defined as follows.
As a toy example, let $\mathcal{S}_1$=\{1, 3, 7, 12\} and $\mathcal{S}_2$=\{2, 3, 7, 9\} be summaries produced by two models, respectively. 
The numbers are sentence indices.
For informativeness, we take the union of summary sentences
\{1, 2, 3, 7, 9, 12\}
and ask human evaluators to judge the relevance of each sentence
against the ground-truth summary
on a scale of 1 (worst) to 5 (best).
The informativeness score of a summary is the average of its sentence scores.
For diversity, we obtain the symmetric difference of two summaries 
\{1, 2, 9, 12\} and ask humans to judge if each sentence has offered new content different from those of the common sentences \{3, 7\}.
A good summary should contain diverse sentences that are dissimilar from each other.

Results are reported in Table~\ref{tab:human_eval}. 
We performed evaluation using Amazon Mechanical Turk on 100 randomly selected summarization instances from arXiv.
arXiv is chosen over other datasets because its content are more accessible to crowd workers. 
We recruited Masters turkers to work on our task.
They must have completed at least 100 HITs and with $\geq$90\% HIT approval rate.
Each summary sentence was judged by 3 turkers. 
Overall, we find that \textbf{Lodoss}-\emph{full} receives better relevancy and diversity ratings than \textbf{Lodoss}-\emph{joint}.
A substantial portion of the sentences receive a score of 1 or 2.
It suggests that the extracted sentences lack informativeness despite that the DPP regularizer is effective at increasing the diversity of selected sentences and eliminating redundancy.

\section{Conclusion}

We tackle the problem of long document extractive summarization
by combining two essential tasks of section segmentation and sentence extraction.
We further design a regularizer drawing on determinantal point processes to 
ensure a set of representative and diverse sentences are selected for the summary.
Extensive experiments and ablations demonstrate the effectiveness of our proposed approach.
Our future work includes exploration of various text segmentation techniques to improve our understanding of the latent document structure. Another direction would be to extend our study to the realm of neural abstractive summarization with the help of learned document structure.

\section{Limitations}

The proposed summarization models are trained on scientific articles that are segmented into multiple sections by authors.
Those section boundaries are utilized by the model to learn robust sentence representations and estimate sentence salience given their proximity to section boundaries.
When section boundaries are unavailable, the model may not work as intended.
Moreover, trained models may carry inductive biases rooted in the data they are pretrained on.
Finetuning on target datasets helps mitigate the issue as 
the model has been shown to demonstrate a reasonable degree of transferability from written documents to other genres.

\section*{Acknowledgements}

We are grateful to the reviewers for their insightful comments that have enriched our paper.
Fei Liu is supported in part by National Science Foundation grant IIS-2303655.

\bibliography{more,fei,summ,abs_summ,trans_summ,anthology}

\begin{thebibliography}{64}
\expandafter\ifx\csname natexlab\endcsname\relax\def\natexlab#1{#1}\fi

\bibitem[{Arnold et~al.(2019)Arnold, Schneider, Cudr{\'e}-Mauroux, Gers, and
  L{\"o}ser}]{arnold-etal-2019-sector}
Sebastian Arnold, Rudolf Schneider, Philippe Cudr{\'e}-Mauroux, Felix~A. Gers,
  and Alexander L{\"o}ser. 2019.
\newblock \href {https://doi.org/10.1162/tacl_a_00261} {{SECTOR}: A neural
  model for coherent topic segmentation and classification}.
\newblock \emph{Transactions of the Association for Computational Linguistics},
  7:169--184.

\bibitem[{Baxendale(1958)}]{Baxendale:1958}
P.~B. Baxendale. 1958.
\newblock \href {https://doi.org/10.1147/rd.24.0354} {Machine-made index for
  technical literature: An experiment}.
\newblock \emph{IBM J. Res. Dev.}, 2(4):354–361.

\bibitem[{Beltagy et~al.(2020)Beltagy, Peters, and Cohan}]{Beltagy:2020}
Iz~Beltagy, Matthew~E. Peters, and Arman Cohan. 2020.
\newblock Longformer: The long-document transformer.
\newblock \emph{arXiv:2004.05150}.

\bibitem[{Carletta et~al.(2006)Carletta, Ashby, Bourban, Flynn, Guillemot,
  Hain, Kadlec, Karaiskos, Kraaij, Kronenthal, Lathoud, Lincoln, Lisowska,
  McCowan, Post, Reidsma, and Wellner}]{Carletta:2006}
Jean Carletta, Simone Ashby, Sebastien Bourban, Mike Flynn, Mael Guillemot,
  Thomas Hain, Jaroslav Kadlec, Vasilis Karaiskos, Wessel Kraaij, Melissa
  Kronenthal, Guillaume Lathoud, Mike Lincoln, Agnes Lisowska, Iain McCowan,
  Wilfried Post, Dennis Reidsma, and Pierre Wellner. 2006.
\newblock The {AMI} meeting corpus: A pre-announcement.
\newblock In \emph{Machine Learning for Multimodal Interaction}, pages 28--39.
  Springer Berlin Heidelberg.

\bibitem[{Chen et~al.(2022)Chen, Chu, Wiseman, and
  Gimpel}]{chen-etal-2022-summscreen}
Mingda Chen, Zewei Chu, Sam Wiseman, and Kevin Gimpel. 2022.
\newblock \href {https://doi.org/10.18653/v1/2022.acl-long.589}
  {{S}umm{S}creen: A dataset for abstractive screenplay summarization}.
\newblock In \emph{Proceedings of the 60th Annual Meeting of the Association
  for Computational Linguistics (Volume 1: Long Papers)}, pages 8602--8615,
  Dublin, Ireland. Association for Computational Linguistics.

\bibitem[{Chen and Bansal(2018)}]{chen-bansal-2018-fast}
Yen-Chun Chen and Mohit Bansal. 2018.
\newblock \href {https://doi.org/10.18653/v1/P18-1063} {Fast abstractive
  summarization with reinforce-selected sentence rewriting}.
\newblock In \emph{Proceedings of the 56th Annual Meeting of the Association
  for Computational Linguistics (Volume 1: Long Papers)}, pages 675--686,
  Melbourne, Australia. Association for Computational Linguistics.

\bibitem[{Child et~al.(2019)Child, Gray, Radford, and
  Sutskever}]{child2019generating}
Rewon Child, Scott Gray, Alec Radford, and Ilya Sutskever. 2019.
\newblock \href {http://arxiv.org/abs/1904.10509} {Generating long sequences
  with sparse transformers}.

\bibitem[{Cho et~al.(2019{\natexlab{a}})Cho, Lebanoff, Foroosh, and
  Liu}]{cho-etal-2019-improving}
Sangwoo Cho, Logan Lebanoff, Hassan Foroosh, and Fei Liu. 2019{\natexlab{a}}.
\newblock \href {https://doi.org/10.18653/v1/P19-1098} {Improving the
  similarity measure of determinantal point processes for extractive
  multi-document summarization}.
\newblock In \emph{Proceedings of the 57th Annual Meeting of the Association
  for Computational Linguistics}, pages 1027--1038, Florence, Italy.
  Association for Computational Linguistics.

\bibitem[{Cho et~al.(2019{\natexlab{b}})Cho, Li, Yu, Foroosh, and
  Liu}]{cho-etal-2019-multi}
Sangwoo Cho, Chen Li, Dong Yu, Hassan Foroosh, and Fei Liu. 2019{\natexlab{b}}.
\newblock \href {https://doi.org/10.18653/v1/D19-5412} {Multi-document
  summarization with determinantal point processes and contextualized
  representations}.
\newblock In \emph{Proceedings of the 2nd Workshop on New Frontiers in
  Summarization}, pages 98--103, Hong Kong, China. Association for
  Computational Linguistics.

\bibitem[{Cohan et~al.(2018)Cohan, Dernoncourt, Kim, Bui, Kim, Chang, and
  Goharian}]{cohan-etal-2018-discourse}
Arman Cohan, Franck Dernoncourt, Doo~Soon Kim, Trung Bui, Seokhwan Kim, Walter
  Chang, and Nazli Goharian. 2018.
\newblock \href {https://doi.org/10.18653/v1/N18-2097} {A discourse-aware
  attention model for abstractive summarization of long documents}.
\newblock In \emph{Proceedings of the 2018 Conference of the North {A}merican
  Chapter of the Association for Computational Linguistics: Human Language
  Technologies, Volume 2 (Short Papers)}, pages 615--621, New Orleans,
  Louisiana. Association for Computational Linguistics.

\bibitem[{Daume~III and Marcu(2002)}]{daume-iii-marcu-2002-noisy}
Hal Daume~III and Daniel Marcu. 2002.
\newblock \href {https://doi.org/10.3115/1073083.1073159} {A noisy-channel
  model for document compression}.
\newblock In \emph{Proceedings of the 40th Annual Meeting of the Association
  for Computational Linguistics}, pages 449--456, Philadelphia, Pennsylvania,
  USA. Association for Computational Linguistics.

\bibitem[{Dror et~al.(2018)Dror, Baumer, Shlomov, and
  Reichart}]{significance_test}
Rotem Dror, Gili Baumer, Segev Shlomov, and Roi Reichart. 2018.
\newblock \href {http://aclweb.org/anthology/P18-1128} {The hitchhiker's guide
  to testing statistical significance in natural language processing}.
\newblock In \emph{Proceedings of the 56th Annual Meeting of the Association
  for Computational Linguistics (Volume 1: Long Papers)}, pages 1383--1392.
  Association for Computational Linguistics.

\bibitem[{Erkan and Radev(2004)}]{Erkan:2004}
G\"{u}nes Erkan and Dragomir~R. Radev. 2004.
\newblock \href {https://www.aaai.org/Papers/JAIR/Vol22/JAIR-2214.pdf}
  {{LexRank}: {G}raph-based lexical centrality as salience in text
  summarization}.
\newblock \emph{Journal of Artificial Intelligence Research}.

\bibitem[{Falcon(2019)}]{falcon2019pytorch}
WA~Falcon. 2019.
\newblock Pytorch lightning.
\newblock \emph{GitHub. Note:
  https://github.com/PyTorchLightning/pytorch-lightning Cited by}, 3.

\bibitem[{Falke et~al.(2019)Falke, Ribeiro, Utama, Dagan, and
  Gurevych}]{falke-etal-2019-ranking}
Tobias Falke, Leonardo F.~R. Ribeiro, Prasetya~Ajie Utama, Ido Dagan, and Iryna
  Gurevych. 2019.
\newblock \href {https://doi.org/10.18653/v1/P19-1213} {Ranking generated
  summaries by correctness: An interesting but challenging application for
  natural language inference}.
\newblock In \emph{Proceedings of the 57th Annual Meeting of the Association
  for Computational Linguistics}, pages 2214--2220, Florence, Italy.
  Association for Computational Linguistics.

\bibitem[{Goldstein et~al.(2000)Goldstein, Mittal, Carbonell, and
  Kantrowitz}]{goldstein-etal-2000-multi}
Jade Goldstein, Vibhu Mittal, Jaime Carbonell, and Mark Kantrowitz. 2000.
\newblock \href {https://www.aclweb.org/anthology/W00-0405} {Multi-document
  summarization by sentence extraction}.
\newblock In \emph{NAACL-ANLP 2000 Workshop: Automatic Summarization}.

\bibitem[{Goyal and Durrett(2021)}]{goyal-durrett-2021-annotating}
Tanya Goyal and Greg Durrett. 2021.
\newblock \href {https://doi.org/10.18653/v1/2021.naacl-main.114} {Annotating
  and modeling fine-grained factuality in summarization}.
\newblock In \emph{Proceedings of the 2021 Conference of the North American
  Chapter of the Association for Computational Linguistics: Human Language
  Technologies}, pages 1449--1462, Online. Association for Computational
  Linguistics.

\bibitem[{Grusky et~al.(2018)Grusky, Naaman, and
  Artzi}]{grusky-etal-2018-newsroom}
Max Grusky, Mor Naaman, and Yoav Artzi. 2018.
\newblock \href {https://doi.org/10.18653/v1/N18-1065} {{N}ewsroom: A dataset
  of 1.3 million summaries with diverse extractive strategies}.
\newblock In \emph{Proceedings of the 2018 Conference of the North {A}merican
  Chapter of the Association for Computational Linguistics: Human Language
  Technologies, Volume 1 (Long Papers)}, pages 708--719, New Orleans,
  Louisiana. Association for Computational Linguistics.

\bibitem[{Hearst(1997)}]{hearst-1997-text}
Marti~A. Hearst. 1997.
\newblock \href {https://www.aclweb.org/anthology/J97-1003} {Text tiling:
  Segmenting text into multi-paragraph subtopic passages}.
\newblock \emph{Computational Linguistics}, 23(1):33--64.

\bibitem[{Huang et~al.(2021)Huang, Cao, Parulian, Ji, and
  Wang}]{huang-etal-2021-efficient}
Luyang Huang, Shuyang Cao, Nikolaus Parulian, Heng Ji, and Lu~Wang. 2021.
\newblock \href {https://doi.org/10.18653/v1/2021.naacl-main.112} {Efficient
  attentions for long document summarization}.
\newblock In \emph{Proceedings of the 2021 Conference of the North American
  Chapter of the Association for Computational Linguistics: Human Language
  Technologies}, pages 1419--1436, Online. Association for Computational
  Linguistics.

\bibitem[{Kedzie et~al.(2018)Kedzie, McKeown, and
  Daum{\'e}~III}]{kedzie-etal-2018-content}
Chris Kedzie, Kathleen McKeown, and Hal Daum{\'e}~III. 2018.
\newblock \href {https://doi.org/10.18653/v1/D18-1208} {Content selection in
  deep learning models of summarization}.
\newblock In \emph{Proceedings of the 2018 Conference on Empirical Methods in
  Natural Language Processing}, pages 1818--1828, Brussels, Belgium.
  Association for Computational Linguistics.

\bibitem[{Koay et~al.(2020)Koay, Roustai, Dai, Burns, Kerrigan, and
  Liu}]{koay-etal-2020-domain}
Jia~Jin Koay, Alexander Roustai, Xiaojin Dai, Dillon Burns, Alec Kerrigan, and
  Fei Liu. 2020.
\newblock \href {https://doi.org/10.18653/v1/2020.coling-main.499} {How domain
  terminology affects meeting summarization performance}.
\newblock In \emph{Proceedings of the 28th International Conference on
  Computational Linguistics}, pages 5689--5695, Barcelona, Spain (Online).
  International Committee on Computational Linguistics.

\bibitem[{Koay et~al.(2021)Koay, Roustai, Dai, and
  Liu}]{koay-etal-2021-sliding}
Jia~Jin Koay, Alexander Roustai, Xiaojin Dai, and Fei Liu. 2021.
\newblock \href {https://doi.org/10.18653/v1/2021.naacl-srw.10} {A
  sliding-window approach to automatic creation of meeting minutes}.
\newblock In \emph{Proceedings of the 2021 Conference of the North American
  Chapter of the Association for Computational Linguistics: Student Research
  Workshop}, pages 68--75, Online. Association for Computational Linguistics.

\bibitem[{Koshorek et~al.(2018)Koshorek, Cohen, Mor, Rotman, and
  Berant}]{koshorek-etal-2018-text}
Omri Koshorek, Adir Cohen, Noam Mor, Michael Rotman, and Jonathan Berant. 2018.
\newblock \href {https://doi.org/10.18653/v1/N18-2075} {Text segmentation as a
  supervised learning task}.
\newblock In \emph{Proceedings of the 2018 Conference of the North {A}merican
  Chapter of the Association for Computational Linguistics: Human Language
  Technologies, Volume 2 (Short Papers)}, pages 469--473, New Orleans,
  Louisiana. Association for Computational Linguistics.

\bibitem[{Kryscinski et~al.(2020)Kryscinski, McCann, Xiong, and
  Socher}]{kryscinski-etal-2020-evaluating}
Wojciech Kryscinski, Bryan McCann, Caiming Xiong, and Richard Socher. 2020.
\newblock \href {https://doi.org/10.18653/v1/2020.emnlp-main.750} {Evaluating
  the factual consistency of abstractive text summarization}.
\newblock In \emph{Proceedings of the 2020 Conference on Empirical Methods in
  Natural Language Processing (EMNLP)}, pages 9332--9346, Online. Association
  for Computational Linguistics.

\bibitem[{Kulesza and Taskar(2012)}]{Kulesza:2012}
Alex Kulesza and Ben Taskar. 2012.
\newblock \href {https://arxiv.org/abs/1207.6083} {\emph{Determinantal Point
  Processes for Machine Learning}}.
\newblock Now Publishers Inc.

\bibitem[{Lebanoff et~al.(2020)Lebanoff, Muchovej, Dernoncourt, Kim, Wang,
  Chang, and Liu}]{lebanoff-etal-2020-understanding}
Logan Lebanoff, John Muchovej, Franck Dernoncourt, Doo~Soon Kim, Lidan Wang,
  Walter Chang, and Fei Liu. 2020.
\newblock \href {https://doi.org/10.18653/v1/2020.acl-srw.26} {Understanding
  points of correspondence between sentences for abstractive summarization}.
\newblock In \emph{Proceedings of the 58th Annual Meeting of the Association
  for Computational Linguistics: Student Research Workshop}, pages 191--198,
  Online. Association for Computational Linguistics.

\bibitem[{Li et~al.(2019)Li, Zhang, Ji, and Radke}]{li-etal-2019-keep}
Manling Li, Lingyu Zhang, Heng Ji, and Richard~J. Radke. 2019.
\newblock \href {https://doi.org/10.18653/v1/P19-1210} {Keep meeting summaries
  on topic: Abstractive multi-modal meeting summarization}.
\newblock In \emph{Proceedings of the 57th Annual Meeting of the Association
  for Computational Linguistics}, pages 2190--2196, Florence, Italy.
  Association for Computational Linguistics.

\bibitem[{Lin(2004)}]{lin-2004-rouge}
Chin-Yew Lin. 2004.
\newblock \href {https://www.aclweb.org/anthology/W04-1013} {{ROUGE}: A package
  for automatic evaluation of summaries}.
\newblock In \emph{Text Summarization Branches Out}, pages 74--81, Barcelona,
  Spain. Association for Computational Linguistics.

\bibitem[{Liu and Lapata(2019)}]{liu-lapata-2019-text}
Yang Liu and Mirella Lapata. 2019.
\newblock \href {https://doi.org/10.18653/v1/D19-1387} {Text summarization with
  pretrained encoders}.
\newblock In \emph{Proceedings of the 2019 Conference on Empirical Methods in
  Natural Language Processing and the 9th International Joint Conference on
  Natural Language Processing (EMNLP-IJCNLP)}, pages 3730--3740, Hong Kong,
  China. Association for Computational Linguistics.

\bibitem[{Liu et~al.(2022)Liu, Liu, Radev, and Neubig}]{liu-etal-2022-brio}
Yixin Liu, Pengfei Liu, Dragomir Radev, and Graham Neubig. 2022.
\newblock \href {https://doi.org/10.18653/v1/2022.acl-long.207} {{BRIO}:
  Bringing order to abstractive summarization}.
\newblock In \emph{Proceedings of the 60th Annual Meeting of the Association
  for Computational Linguistics (Volume 1: Long Papers)}, pages 2890--2903,
  Dublin, Ireland. Association for Computational Linguistics.

\bibitem[{Lukasik et~al.(2020)Lukasik, Dadachev, Papineni, and
  Sim{\~o}es}]{lukasik-etal-2020-text}
Michal Lukasik, Boris Dadachev, Kishore Papineni, and Gon{\c{c}}alo Sim{\~o}es.
  2020.
\newblock \href {https://doi.org/10.18653/v1/2020.emnlp-main.380} {Text
  segmentation by cross segment attention}.
\newblock In \emph{Proceedings of the 2020 Conference on Empirical Methods in
  Natural Language Processing (EMNLP)}, pages 4707--4716, Online. Association
  for Computational Linguistics.

\bibitem[{Lv et~al.(2021)Lv, Cui, Vasilijevic, and Wei}]{lv2021vt}
Tengchao Lv, Lei Cui, Momcilo Vasilijevic, and Furu Wei. 2021.
\newblock Vt-ssum: A benchmark dataset for video transcript segmentation and
  summarization.
\newblock \emph{arXiv preprint arXiv:2106.05606}.

\bibitem[{Lyons(2021)}]{HighlightedLinks:2021}
Kim Lyons. 2021.
\newblock \href
  {https://www.theverge.com/2021/4/17/22389519/google-feature-chrome-90-highlighted-links}
  {Google introducing a feature in chrome 90 to create links to highlighted
  text on a webpage}.
\newblock \emph{The Verge}.

\bibitem[{Malioutov and Barzilay(2006)}]{malioutov-barzilay-2006-minimum}
Igor Malioutov and Regina Barzilay. 2006.
\newblock \href {https://doi.org/10.3115/1220175.1220179} {Minimum cut model
  for spoken lecture segmentation}.
\newblock In \emph{Proceedings of the 21st International Conference on
  Computational Linguistics and 44th Annual Meeting of the Association for
  Computational Linguistics}, pages 25--32, Sydney, Australia. Association for
  Computational Linguistics.

\bibitem[{Mao et~al.(2021)Mao, Wu, Ni, Zhang, Zhang, Yu, Deb, Zhu, Awadallah,
  and Radev}]{mao2021dyle}
Ziming Mao, Chen~Henry Wu, Ansong Ni, Yusen Zhang, Rui Zhang, Tao Yu,
  Budhaditya Deb, Chenguang Zhu, Ahmed~H. Awadallah, and Dragomir Radev. 2021.
\newblock \href {http://arxiv.org/abs/2110.08168} {Dyle: Dynamic latent
  extraction for abstractive long-input summarization}.

\bibitem[{Marcu(1998)}]{marcu-1998-improving}
Daniel Marcu. 1998.
\newblock \href {https://www.aclweb.org/anthology/W98-1124} {Improving
  summarization through rhetorical parsing tuning}.
\newblock In \emph{Sixth Workshop on Very Large Corpora}.

\bibitem[{Maynez et~al.(2020)Maynez, Narayan, Bohnet, and
  McDonald}]{maynez-etal-2020-faithfulness}
Joshua Maynez, Shashi Narayan, Bernd Bohnet, and Ryan McDonald. 2020.
\newblock \href {https://doi.org/10.18653/v1/2020.acl-main.173} {On
  faithfulness and factuality in abstractive summarization}.
\newblock In \emph{Proceedings of the 58th Annual Meeting of the Association
  for Computational Linguistics}, pages 1906--1919, Online. Association for
  Computational Linguistics.

\bibitem[{Narayan et~al.(2018)Narayan, Cohen, and
  Lapata}]{narayan-etal-2018-ranking}
Shashi Narayan, Shay~B. Cohen, and Mirella Lapata. 2018.
\newblock \href {https://doi.org/10.18653/v1/N18-1158} {Ranking sentences for
  extractive summarization with reinforcement learning}.
\newblock In \emph{Proceedings of the 2018 Conference of the North {A}merican
  Chapter of the Association for Computational Linguistics: Human Language
  Technologies, Volume 1 (Long Papers)}, pages 1747--1759, New Orleans,
  Louisiana. Association for Computational Linguistics.

\bibitem[{Narayan et~al.(2020)Narayan, Maynez, Adamek, Pighin, Bratanic, and
  McDonald}]{narayan-etal-2020-stepwise}
Shashi Narayan, Joshua Maynez, Jakub Adamek, Daniele Pighin, Blaz Bratanic, and
  Ryan McDonald. 2020.
\newblock \href {https://doi.org/10.18653/v1/2020.emnlp-main.339} {Stepwise
  extractive summarization and planning with structured transformers}.
\newblock In \emph{Proceedings of the 2020 Conference on Empirical Methods in
  Natural Language Processing (EMNLP)}, pages 4143--4159, Online. Association
  for Computational Linguistics.

\bibitem[{Pagnoni et~al.(2021)Pagnoni, Balachandran, and
  Tsvetkov}]{pagnoni-etal-2021-understanding}
Artidoro Pagnoni, Vidhisha Balachandran, and Yulia Tsvetkov. 2021.
\newblock \href {https://doi.org/10.18653/v1/2021.naacl-main.383}
  {Understanding factuality in abstractive summarization with {FRANK}: A
  benchmark for factuality metrics}.
\newblock In \emph{Proceedings of the 2021 Conference of the North American
  Chapter of the Association for Computational Linguistics: Human Language
  Technologies}, pages 4812--4829, Online. Association for Computational
  Linguistics.

\bibitem[{Passonneau and Litman(1997)}]{passonneau-litman-1997-discourse}
Rebecca~J. Passonneau and Diane~J. Litman. 1997.
\newblock \href {https://www.aclweb.org/anthology/J97-1005} {Discourse
  segmentation by human and automated means}.
\newblock \emph{Computational Linguistics}, 23(1):103--139.

\bibitem[{Paszke et~al.(2019)Paszke, Gross, Massa, Lerer, Bradbury, Chanan,
  Killeen, Lin, Gimelshein, Antiga, Desmaison, Kopf, Yang, DeVito, Raison,
  Tejani, Chilamkurthy, Steiner, Fang, Bai, and
  Chintala}]{PyTorch:NEURIPS2019_9015}
Adam Paszke, Sam Gross, Francisco Massa, Adam Lerer, James Bradbury, Gregory
  Chanan, Trevor Killeen, Zeming Lin, Natalia Gimelshein, Luca Antiga, Alban
  Desmaison, Andreas Kopf, Edward Yang, Zachary DeVito, Martin Raison, Alykhan
  Tejani, Sasank Chilamkurthy, Benoit Steiner, Lu~Fang, Junjie Bai, and Soumith
  Chintala. 2019.
\newblock \href
  {http://papers.neurips.cc/paper/9015-pytorch-an-imperative-style-high-performance-deep-learning-library.pdf}
  {Pytorch: An imperative style, high-performance deep learning library}.
\newblock In H.~Wallach, H.~Larochelle, A.~Beygelzimer, F.~d\textquotesingle
  Alch\'{e}-Buc, E.~Fox, and R.~Garnett, editors, \emph{Advances in Neural
  Information Processing Systems 32}, pages 8024--8035. Curran Associates, Inc.

\bibitem[{Perez-Beltrachini and Lapata(2021)}]{DPPAttn:2021}
Laura Perez-Beltrachini and Mirella Lapata. 2021.
\newblock \href {https://doi.org/10.1613/jair.1.12522} {Multi-document
  summarization with determinantal point process attention}.
\newblock \emph{J. Artif. Int. Res.}, 71:371–399.

\bibitem[{Pevzner and Hearst(2002)}]{pevzner-hearst-2002-critique}
Lev Pevzner and Marti~A. Hearst. 2002.
\newblock \href {https://doi.org/10.1162/089120102317341756} {A critique and
  improvement of an evaluation metric for text segmentation}.
\newblock \emph{Computational Linguistics}, 28(1):19--36.

\bibitem[{Pilault et~al.(2020)Pilault, Li, Subramanian, and
  Pal}]{pilault-etal-2020-extractive}
Jonathan Pilault, Raymond Li, Sandeep Subramanian, and Chris Pal. 2020.
\newblock \href {https://doi.org/10.18653/v1/2020.emnlp-main.748} {On
  extractive and abstractive neural document summarization with transformer
  language models}.
\newblock In \emph{Proceedings of the 2020 Conference on Empirical Methods in
  Natural Language Processing (EMNLP)}, pages 9308--9319, Online. Association
  for Computational Linguistics.

\bibitem[{Power et~al.(2003)Power, Scott, and
  Bouayad-Agha}]{power-etal-2003-document}
Richard Power, Donia Scott, and Nadjet Bouayad-Agha. 2003.
\newblock \href {https://doi.org/10.1162/089120103322145315} {Document
  structure}.
\newblock \emph{Computational Linguistics}, 29(2):211--260.

\bibitem[{Riezler and Maxwell(2005)}]{riezler-maxwell-2005-pitfalls}
Stefan Riezler and John~T. Maxwell. 2005.
\newblock \href {https://www.aclweb.org/anthology/W05-0908} {On some pitfalls
  in automatic evaluation and significance testing for {MT}}.
\newblock In \emph{Proceedings of the {ACL} Workshop on Intrinsic and Extrinsic
  Evaluation Measures for Machine Translation and/or Summarization}, pages
  57--64, Ann Arbor, Michigan. Association for Computational Linguistics.

\bibitem[{Rohde et~al.(2021)Rohde, Wu, and Liu}]{Rohde-etal_2020}
Tobias Rohde, Xiaoxia Wu, and Yinhan Liu. 2021.
\newblock \href {http://arxiv.org/abs/2104.07545} {Hierarchical learning for
  generation with long source sequences}.
\newblock \emph{CoRR}, abs/2104.07545.

\bibitem[{Roy et~al.(2021)Roy, Saffar, Vaswani, and
  Grangier}]{roy-etal-2021-efficient}
Aurko Roy, Mohammad Saffar, Ashish Vaswani, and David Grangier. 2021.
\newblock \href {https://doi.org/10.1162/tacl_a_00353} {Efficient content-based
  sparse attention with routing transformers}.
\newblock \emph{Transactions of the Association for Computational Linguistics},
  9:53--68.

\bibitem[{See et~al.(2017)See, Liu, and Manning}]{see-etal-2017-get}
Abigail See, Peter~J. Liu, and Christopher~D. Manning. 2017.
\newblock \href {https://doi.org/10.18653/v1/P17-1099} {Get to the point:
  Summarization with pointer-generator networks}.
\newblock In \emph{Proceedings of the 55th Annual Meeting of the Association
  for Computational Linguistics (Volume 1: Long Papers)}, pages 1073--1083,
  Vancouver, Canada. Association for Computational Linguistics.

\bibitem[{Sennrich et~al.(2016)Sennrich, Haddow, and
  Birch}]{sennrich-etal-2016-neural}
Rico Sennrich, Barry Haddow, and Alexandra Birch. 2016.
\newblock \href {https://doi.org/10.18653/v1/P16-1162} {Neural machine
  translation of rare words with subword units}.
\newblock In \emph{Proceedings of the 54th Annual Meeting of the Association
  for Computational Linguistics (Volume 1: Long Papers)}, pages 1715--1725,
  Berlin, Germany. Association for Computational Linguistics.

\bibitem[{Shang et~al.(2018)Shang, Ding, Zhang, Tixier, Meladianos,
  Vazirgiannis, and Lorr{\'e}}]{shang-etal-2018-unsupervised}
Guokan Shang, Wensi Ding, Zekun Zhang, Antoine Tixier, Polykarpos Meladianos,
  Michalis Vazirgiannis, and Jean-Pierre Lorr{\'e}. 2018.
\newblock \href {https://doi.org/10.18653/v1/P18-1062} {Unsupervised
  abstractive meeting summarization with multi-sentence compression and
  budgeted submodular maximization}.
\newblock In \emph{Proceedings of the 56th Annual Meeting of the Association
  for Computational Linguistics (Volume 1: Long Papers)}, pages 664--674,
  Melbourne, Australia. Association for Computational Linguistics.

\bibitem[{Shriberg(1994)}]{Shriberg:1994}
Elizabeth Shriberg. 1994.
\newblock Preliminaries to a theory of speech disfluencies.
\newblock \emph{Ph.D. thesis, Department of Psychology, University of
  California, Berkeley}.

\bibitem[{Vanderwende et~al.(2007)Vanderwende, Suzuki, Brockett, and
  Nenkova}]{Vanderwende:2007}
Lucy Vanderwende, Hisami Suzuki, Chris Brockett, and Ani Nenkova. 2007.
\newblock \href {https://www.cis.upenn.edu/~nenkova/papers/ipm.pdf} {Beyond
  {SumBasic}: {T}ask-focused summarization with sentence simplification and
  lexical expansion}.
\newblock \emph{Information Processing and Management}, 43(6):1606--1618.

\bibitem[{Wolf et~al.(2020)Wolf, Debut, Sanh, Chaumond, Delangue, Moi, Cistac,
  Rault, Louf, Funtowicz, Davison, Shleifer, von Platen, Ma, Jernite, Plu, Xu,
  Scao, Gugger, Drame, Lhoest, and Rush}]{wolf-etal-2020-transformers}
Thomas Wolf, Lysandre Debut, Victor Sanh, Julien Chaumond, Clement Delangue,
  Anthony Moi, Pierric Cistac, Tim Rault, Rémi Louf, Morgan Funtowicz, Joe
  Davison, Sam Shleifer, Patrick von Platen, Clara Ma, Yacine Jernite, Julien
  Plu, Canwen Xu, Teven~Le Scao, Sylvain Gugger, Mariama Drame, Quentin Lhoest,
  and Alexander~M. Rush. 2020.
\newblock \href {https://www.aclweb.org/anthology/2020.emnlp-demos.6}
  {Transformers: State-of-the-art natural language processing}.
\newblock In \emph{Proceedings of the 2020 Conference on Empirical Methods in
  Natural Language Processing: System Demonstrations}, pages 38--45, Online.
  Association for Computational Linguistics.

\bibitem[{Xiao and Carenini(2019)}]{xiao-carenini-2019-extractive}
Wen Xiao and Giuseppe Carenini. 2019.
\newblock \href {https://doi.org/10.18653/v1/D19-1298} {Extractive
  summarization of long documents by combining global and local context}.
\newblock In \emph{Proceedings of the 2019 Conference on Empirical Methods in
  Natural Language Processing and the 9th International Joint Conference on
  Natural Language Processing (EMNLP-IJCNLP)}, pages 3011--3021, Hong Kong,
  China. Association for Computational Linguistics.

\bibitem[{Xiao and Carenini(2020)}]{xiao-carenini-2020-systematically}
Wen Xiao and Giuseppe Carenini. 2020.
\newblock \href {https://aclanthology.org/2020.aacl-main.51} {Systematically
  exploring redundancy reduction in summarizing long documents}.
\newblock In \emph{Proceedings of the 1st Conference of the Asia-Pacific
  Chapter of the Association for Computational Linguistics and the 10th
  International Joint Conference on Natural Language Processing}, pages
  516--528, Suzhou, China. Association for Computational Linguistics.

\bibitem[{Xing et~al.(2020)Xing, Hackinen, Carenini, and
  Trebbi}]{xing-etal-2020-improving}
Linzi Xing, Brad Hackinen, Giuseppe Carenini, and Francesco Trebbi. 2020.
\newblock \href {https://aclanthology.org/2020.aacl-main.63} {Improving context
  modeling in neural topic segmentation}.
\newblock In \emph{Proceedings of the 1st Conference of the Asia-Pacific
  Chapter of the Association for Computational Linguistics and the 10th
  International Joint Conference on Natural Language Processing}, pages
  626--636, Suzhou, China. Association for Computational Linguistics.

\bibitem[{Zaheer et~al.(2020)Zaheer, Guruganesh, Dubey, Ainslie, Alberti,
  Ontanon, Pham, Ravula, Wang, Yang et~al.}]{zaheer2020bigbird}
Manzil Zaheer, Guru Guruganesh, Kumar~Avinava Dubey, Joshua Ainslie, Chris
  Alberti, Santiago Ontanon, Philip Pham, Anirudh Ravula, Qifan Wang, Li~Yang,
  et~al. 2020.
\newblock Big bird: Transformers for longer sequences.
\newblock \emph{Advances in Neural Information Processing Systems}, 33.

\bibitem[{Zhang et~al.(2016)Zhang, Chao, Sha, and Grauman}]{Zhang:2016:DPP}
Ke~Zhang, Wei-Lun Chao, Fei Sha, and Kristen Grauman. 2016.
\newblock \href {https://arxiv.org/abs/1605.08110} {Video summarization with
  long short-term memory}.
\newblock In \emph{Proceedings of the European Conference on Computer Vision
  (ECCV)}.

\bibitem[{Zhang et~al.(2019)Zhang, Wei, and Zhou}]{zhang-etal-2019-hibert}
Xingxing Zhang, Furu Wei, and Ming Zhou. 2019.
\newblock \href {https://doi.org/10.18653/v1/P19-1499} {{HIBERT}: Document
  level pre-training of hierarchical bidirectional transformers for document
  summarization}.
\newblock In \emph{Proceedings of the 57th Annual Meeting of the Association
  for Computational Linguistics}, pages 5059--5069, Florence, Italy.
  Association for Computational Linguistics.

\bibitem[{Zhong et~al.(2021)Zhong, Yin, Yu, Zaidi, Mutuma, Jha, Awadallah,
  Celikyilmaz, Liu, Qiu, and Radev}]{zhong-etal-2021-qmsum}
Ming Zhong, Da~Yin, Tao Yu, Ahmad Zaidi, Mutethia Mutuma, Rahul Jha,
  Ahmed~Hassan Awadallah, Asli Celikyilmaz, Yang Liu, Xipeng Qiu, and Dragomir
  Radev. 2021.
\newblock \href {https://doi.org/10.18653/v1/2021.naacl-main.472} {{QMS}um: A
  new benchmark for query-based multi-domain meeting summarization}.
\newblock In \emph{Proceedings of the 2021 Conference of the North American
  Chapter of the Association for Computational Linguistics: Human Language
  Technologies}, pages 5905--5921, Online. Association for Computational
  Linguistics.

\bibitem[{Zhu et~al.(2020)Zhu, Xu, Zeng, and
  Huang}]{zhu-etal-2020-hierarchical}
Chenguang Zhu, Ruochen Xu, Michael Zeng, and Xuedong Huang. 2020.
\newblock \href {https://doi.org/10.18653/v1/2020.findings-emnlp.19} {A
  hierarchical network for abstractive meeting summarization with cross-domain
  pretraining}.
\newblock In \emph{Findings of the Association for Computational Linguistics:
  EMNLP 2020}, pages 194--203, Online. Association for Computational
  Linguistics.

\end{thebibliography}
\bibliographystyle{acl_natbib}


\end{document}